\documentclass{elsarticle}

\usepackage{float}

\usepackage{amsthm, amsmath, amsfonts}
\usepackage{tikz}
\usepackage{pgfplots} 
\pgfplotsset{compat=1.11}
\usepackage{subcaption}
\usepackage{booktabs}
\usepackage{changes} 
\usepackage[ruled,vlined]{algorithm2e} 
\SetCommentSty{mycommfont} 
\usepackage{adjustbox} 
\usepackage{multirow} 

\makeatletter \renewcommand*{\fps@figure}{h} \makeatother 

\newcommand{\Span}{\text{Span}}
\DeclareMathOperator{\Rips}{Rips}
\DeclareMathOperator*{\kmax}{kmax}
\newcommand*{\defeq}{\mathrel{\vcenter{\baselineskip0.5ex\lineskiplimit0pt\hbox{\scriptsize.}\hbox{\scriptsize.}}}=}
\newcommand{\norm}[1]{\left\lVert#1\right\rVert}
\newcommand{\highest}[1]{{$\mathbf{#1}$}}
\newcommand{\fs}{\textit{fs}}

\graphicspath{{fastFigures/}} 

\biboptions{authoryear}

\renewcommand{\cite}{\citep}

\makeatletter
\def\ps@pprintTitle{%
 \let\@oddhead\@empty
 \let\@evenhead\@empty
 \def\@oddfoot{}%
 \let\@evenfoot\@oddfoot}
\makeatother

\begin{document}

\begin{frontmatter}

\title{Time Series Classification via Topological Data Analysis}

\author[1]{Alperen Karan\corref{cor1}}
\ead{karana@itu.edu.tr.}
\author[1]{Atabey Kaygun}
\ead{kaygun@itu.edu.tr}

\cortext[cor1]{Corresponding author}
\address[1]{Department of Mathematical Engineering, Istanbul Technical University, 34467, Istanbul, Turkey}

\begin{abstract}
In this paper, we develop topological data analysis methods for classification tasks on univariate time series. As an application, we perform binary and ternary classification tasks on two public datasets that consist of physiological signals collected under stress and non-stress conditions. We accomplish our goal by using persistent homology to engineer stable topological features after we use a time delay embedding of the signals and perform a subwindowing instead of using windows of fixed length. The combination of methods we use can be applied to any univariate time series and allows us to reduce noise and use long window sizes without incurring an extra computational cost. We then use machine learning models on the features we algorithmically engineered to obtain higher accuracies with fewer features.
\end{abstract}


\end{frontmatter}

\section{Introduction}

In this study, we use persistent homology to perform classification tasks on two publicly available multivariate time series datasets \cite{schmidt2018introducing, healey2005detecting} that include physiological data collected during stressful and non-stressful tasks. {To make our analyses more thorough, we also tested our methods on a synthetic time series dataset.} Instead of directly computing signal-specific features from \emph{sliding windows}  and \emph{subwindows} on modalities such as electrocardiogram and wrist temperature (Figure \ref{subwindowing_method}), we extracted features using \emph{persistence diagrams} and their statistical properties. Subwindowing method allowed us to reduce noise without incurring an extra computational cost. We then developed machine learning models and then assessed the performance of our models by varying window sizes and using different flavors of persistence diagrams.

Topological Data Analysis (TDA) techniques usually work with points embedded in an affine space of large enough dimension. However, TDA techniques can still be applied to time series data sets, whether they are univariate or multivariate. One can convert a univariate time series into a finite collection of points in a $d$-dimensional affine space using \emph{delay embedding} methods, of which one can compute persistent homology. Since Taken's Theorem implies that the delay embeddings produce topologically invariant subsets on a non-chaotical dynamical system \cite{takens1981detecting}, one can reasonably expect that persistent homology produces features that would distinguish different time series.

There is a handful of research for time series classification using topological data analysis. The reader may refer to \citet{perea2019topological} for theoretical background and real-world applications of sliding window persistence. A typical strategy for such classification tasks is to create equally sized sliding windows from the time series, then to find the corresponding delay embeddings and compute the persistent homology. If the windows are long, then the resulting dataset becomes large, and thus, persistent homology computation becomes prohibitive. In such cases, subsampling can be used to reduce the dataset size \cite[e.g.,][]{dirafzoon2016action, emrani2014persistent}, or the time series can be divided into non-overlapping subseries which are then converted to delay embeddings and to multiple persistence diagrams \cite{majumdar2020clustering}. Persistent homology computation can also be made on the level set filtration rather than delay embeddings \cite{majumder2020detecting} or from both \cite{chung2021persistent}.

Once a persistence diagram for each window is obtained, distance-based machine learning models (such as kNN) can be employed~\cite[e.g.,][]{seversky2016time, marchese2018signal}. However, since finding distance between persistence diagrams is computationally expensive, such machine learning models can only be used if the number of windows remains fairly low. Otherwise, one must resort to secondary methods to extract features from persistence diagrams first. In such cases, one must first engineer feature vectors and then use machine learning models on these feature vectors. 

There are studies that extract features from statistical properties of the diagrams, such as the mean and standard deviations of lifetimes \cite{ignacio2019classification, chung2021persistent}, but such features are sensitive to noise and should be used very cautiously. One can also map a persistence diagram to a function such as a Betti curve or a Persistence Landscape. As these functions lie in a vector space, they can directly be fed into learning algorithms \cite[e.g.,][]{umeda2017time, majumdar2020clustering}, or one can engineer features (such as $L^1$ norm) on these curves \cite{wang2019statistical}. More stable features such as persistent entropy and maximum persistence are also widely used in many applications \cite{majumder2020detecting, emrani2014persistent}. We refer the reader to \citet{pun2018persistent} for a survey of different feature engineering techniques using persistent homology. 

Studies using sliding window persistence have a variety of applications such as action classification \cite{dirafzoon2016action}, wheeze detection \cite{emrani2014persistent}, chatter detection \cite{khasawneh2016chatter}, (quasi)periodicity quantification of videos \cite{tralie2018quasi}, chaos detection \cite{tempelman2020look} and financial time series classification \cite{majumdar2020clustering}. There are also many studies that employ persistent homology of physiological signals for a range of applications, including activity classification from EMG \cite{umeda2017time}, detecting autism spectrum disorder from EEG \cite{majumder2020detecting}, optimal delay embedding parameter selection for EEG \cite{altindics2021parameter}, classification of ECG \cite{ignacio2019classification}, and respiration rate estimation \cite{erden2017period}.

In our study, we start with using sliding windows of a fixed size since this is going to allow us to compare our findings with the original studies~\cite{schmidt2018introducing, healey2005detecting}. Moreover, within each window, we run another sliding window (which we will call a \emph{subwindow}) with a much shorter length, yet long enough to contain at least one cycle of periodic-like signals. Thus the subwindows capture the local information about the window they were taken from, and one can measure statistically how these local features vary over the window.  We specifically created different delay embeddings with different embedding sizes, and then constructed topologically stable features from the resulting persistence diagrams. Finally, we trained and tested machine learning models on these newly engineered features.

\subsection{Plan of the article}

The rest of this paper is set up as follows. In Section \ref{section:tda}, we start by reviewing persistent homology, metrics and kernels on the space of persistence diagrams. {In Section \ref{sect:featureEngineering}, we show the feature engineering methods} on persistence diagrams using the metrics and kernels defined. In Section \ref{section_sliding_windows}, we investigate sliding windows and time delay embeddings of univariate time series, and we demonstrate how feature engineering on sliding subwindows is independent of any window size. In Section \ref{sect:dataset_description}, we discuss the datasets we use in this study.  We explain our methodology in detail in Section \ref{sect:methodology}. We then present our results and their analysis in Section \ref{sect:results} and Section \ref{sect:conclusion}. Finally, in {Section \ref{appendix_section}, we provide the pseudocodes for the algorithms used in this study.}

\section{Topological Data Analysis and Persistent Homology}
\label{section:tda}

{Most of the material we present in this section is used to set up the background and the notation, and can be found in most books and articles on TDA. We refer the reader to \citet{edelsbrunner2010computational} for a detailed introduction on the subject.}

We work within the metric space $\mathbb{R}^n$ with a fixed distance function $d$ and a fixed dimension $n\geq 1$. The reader might assume $d$ is the euclidean metric even though the arguments we provide work equally well with any other metric. We use a finite sample of points $X$ from an unknown region $\Omega$ in $\mathbb{R}^n$. Our ultimate aim is to gain some insight into topological/homological invariants of the region $\Omega$ using only $X$.

\subsection{Simplices}

Given a finite set of points $X$ in $\mathbb{R}^n$, the simplex $S(X)$ spanned by $X$ is the convex hull spanned by $X$.
The points in $X$ are called \emph{the vertices} of the simplex $S(X)$. 
Any subset $Y\subseteq X$ spans a subcomplex $S(Y)$ 
called \emph{faces} of the complex $S(X)$.  The dimension of the face, and therefore the simplicial complex itself, is determined by the number of points. 

\subsection{Simplicial complexes}

An $m$-dimensional \emph{simplicial complex} $C_* = (C_0,\ldots,C_m)$ in $\mathbb{R}^n$ is a finite graded collection of simplices where simplices with a fixed dimension-$i$ is denoted by $C_i$. 
We also ask that given any two simplices $S$ and $S'$ in $C_*$, the intersection $S\cap S'$ is a subsimplex of both $S$ and $S'$. 
The dimension of a simplicial complex $C_*$ is the maximal dimension among all of the simplicial contained in $C_*$.


\subsection{Homology of a complex}

Given a simplicial complex $C_* = C_1,\ldots,C_m$ in $\mathbb{R}^n$, we form a $\mathbb{R}$-vector space $\Span(C_i)$ spanned by $i$-dimensional simplices of $C_*$ together with operators $d_i\colon \Span(C_i)\to \Span(C_{i-1})$ defined as 
\begin{equation}
  \label{eq:1}
   d_i[\mathbf{x}_1,\ldots,\mathbf{x}_i] = \sum_{j=1}^i (-1)^j
  [\mathbf{x}_1,\ldots,\widehat{\mathbf{x}_j},\ldots,\mathbf{x}_i]
\end{equation}
where $\widehat{\mathbf{x}}_j$ indicates that specific vertex is missing.  Now, $d_{i-1}d_i = 0$ for every $i>1$, and therefore, $ker(d_i)\supseteq im(d_{i+1})$. Then we define the $i$-the homology $H_i(C_*)$ as the quotient subspace $ker(d_i)/im(d_{i+1})$ for every $i\geq 0$.  For each $i\geq 0$, the number $\dim_{\mathbb{R}} H_i(C_*)$ is called the \emph{$i$-th Betti number of the complex $C_*$}.
  
\subsection{Rips complexes}\label{subsect:Rips}

One of the most commonly used simplicial complexes obtained from a finite sample of points is the \emph{Rips complex.}  Given $ X$ and $\varepsilon > 0$, a finite set of points $U\subset  X$ forms a simplex if every pair of points in $\mathbf{x},\mathbf{y}\in U$ are at-most $\varepsilon$-apart. The resulting simplicial complex is denoted as $\Rips(X,\varepsilon)$.  
For $\varepsilon = 0$, the Rips complex consists only of the sampled vertices $ X$, which is our the data set, and for a large $\varepsilon$, the Rips complex becomes a single high dimensional simplex.

\subsection{Filtrations}

A filtered complex is a collection of complexes $F_{\varepsilon}$ such that $F_{\varepsilon_1}\subseteq F_{\varepsilon_2}$ whenever $\varepsilon_1\leq \varepsilon_2$. The Rips complex we defined in Subsection \ref{subsect:Rips} is an example. 
One can compute the homology of a Rips complex at a particular $\varepsilon_0$. However, a choice of $\varepsilon_0$ at which the Rips complex has the homology type of $\Omega$ may not necessarily exist unless $ X\subseteq \Omega$ is $\varepsilon$-dense for some $\varepsilon$\footnote{A subset $ X\subseteq \Omega$ is called $\varepsilon$-dense if for every $\mathbf{x}\in \Omega$ there is a point $\mathbf{y}\in X$ such that $d(\mathbf{x},\mathbf{y})<\varepsilon$.  See ~\cite{Bjorner1995} and \cite[page 10]{OtterEtAl2017}.}. So, we need to look at the ``persistent'' features of the homology of the data at different scales of $\varepsilon$ in order to see which homology classes live for a small range of $\varepsilon$, and which homology classes persist longer.

\begin{figure}
\centering
\input{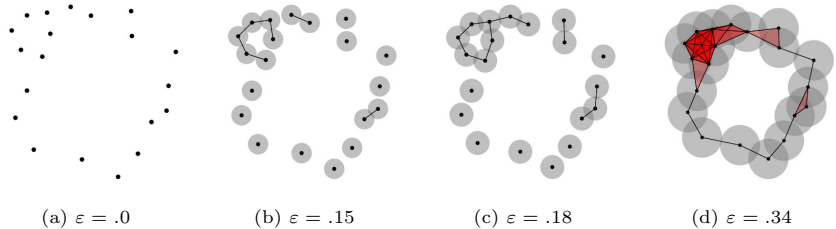}
\caption{The Rips complex $\Rips( X,2\varepsilon)$ (black and red) of a sample dataset $ X$ (grey) drawn at different scales. At $\varepsilon = .18$, there are $10$ components and one hole. At $\varepsilon = .34$, there is one component and one hole.}
\label{filtration}
\end{figure}

\subsection{Persistent homology}

Consider a finite filtration $(K_\varepsilon)_{\varepsilon \in I}$ of simplicial complexes. 
By definition, $\varepsilon_i \leq \varepsilon_j$ implies $K_{\varepsilon_i} \subseteq K_{\varepsilon_j}$. This relation yields a linear map between the homology groups $H_n(K_{\varepsilon_i}) \to H_n(K_{\varepsilon_j})$ for arbitrary $n$~\cite{chazal2012structure}. 
Moreover, there is an interval $[b,d]$ where $b$ is the minimum index where a $H_n(K_{\varepsilon_b})$ is non-zero, and $d$ is the minimum index after which $H_n(K_{\varepsilon_\ell})$ are all 0. The numbers $b$ and $d$ are called \emph{the birth time} and \emph{the death time} of the corresponding homological features.

The multiset of pairs $(b_i,d_i)$ of birth and death times is called the \emph{persistence diagram} \cite{edelsbrunner2000topological}. 
A persistence diagram is assumed to contain infinitely many copies of the \emph{diagonal} which includes the points with simultaneous birth and death. This allows us to create bijections between two persistence diagrams.

\subsection{Wasserstein and Bottleneck distances}

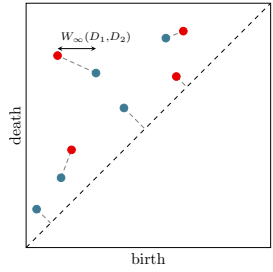
\begin{figure}
\centering
\begin{tikzpicture}[scale = .6]
\begin{axis}[width=7cm, height=7cm, ylabel=death, xlabel=birth, ymin=0, ymax=7, xmin=0, xmax=7, ticks=none]
\draw[dashed, black] (0,0) -- (7,7);

\draw[<->, black, >=stealth] (.9,5.7) -- (2,5.7) node [above] {$\scriptstyle W_\infty(D_1,D_2)$};

\draw[densely dashed, gray] (1,2) -- (1.3,2.8);
\draw[densely dashed, gray] (2,5) -- (.9,5.5);
\draw[densely dashed, gray] (4,6) -- (4.5,6.2);

\draw[densely dashed, gray] (2.8,4) -- (3.4,3.4);
\draw[densely dashed, gray] (.3,1.1) -- (.7,.7);
\draw[densely dashed, gray] (4.3,4.9) -- (4.6,4.6);

\draw[cyan!50!black, fill] (1,2) circle[radius=2.4pt];
\draw[red!90!black, fill] (1.3,2.8) circle[radius=2.4pt];

\draw[cyan!50!black, fill] (2,5) circle[radius=2.4pt];
\draw[red!90!black, fill] (.9,5.5) circle[radius=2.4pt];

\draw[cyan!50!black, fill] (4,6) circle[radius=2.4pt];
\draw[red!90!black, fill] (4.5,6.2) circle[radius=2.4pt];

\draw[cyan!50!black, fill] (2.8,4) circle[radius=2.4pt];
\draw[cyan!50!black, fill] (.3,1.1) circle[radius=2.4pt];
\draw[red!90!black, fill] (4.3,4.9) circle[radius=2.4pt];

\end{axis}
\end{tikzpicture}
\caption{The bottleneck distance between two persistence diagrams.}
\label{bottleneckDistance}
\end{figure}

Let $D_1, D_2$ be persistence diagrams along with their diagonals, and let $\gamma: D_1 \to D_2$ be a bijection. The \emph{bottleneck distance} $W_\infty$ \cite{cohen2007stability} between $D_1$ and $D_2$ is defined as follows (see Figure \ref{bottleneckDistance}): \[ W_\infty(D_1,D_2) \defeq \inf_\gamma \sup_{x \in D_1} \norm{x-\gamma(x)}_\infty. \]
The $p-$\emph{Wasserstein distance} \cite{cohen2010lipschitz} is defined similarly:
\[ W_p(D_1,D_2) \defeq \inf_\gamma \left(\sum_{x \in D_1} \norm{x-\gamma(x)}_\infty^p\right)^{\frac 1p}. \]
The Bottleneck distance is stable under small perturbations of the data, and the $p-$Wasserstein distance is stable under certain assumptions.

\subsection{Persistence landscapes and Betti curves}

\begin{figure}
\centering
\begin{subfigure}{.24\textwidth}\centering
\begin{tikzpicture}[scale=.67]
\begin{axis}[width=5cm, height=5cm, ylabel=death, xlabel=birth, ymin=0, ymax=7, xmin=0, xmax=7, xtick={1,2,3,4,5,6}, ytick={1,2,3,4,5,6}]
\draw[dashed, black] (0,0) -- (7,7);

\draw[dashed, gray] (1,0) -- (1,2) -- (0,2);
\draw[dashed, gray] (2,0) -- (2,5) -- (0,5);
\draw[dashed, gray] (4,0) -- (4,6) -- (0,6);

\draw[cyan!70!black, fill] (1,2) circle[radius=2pt];
\draw[cyan!70!black, fill] (2,5) circle[radius=2pt];
\draw[cyan!70!black, fill] (4,6) circle[radius=2pt];
\end{axis}
\end{tikzpicture}
\caption{Persistence Diagram} 
\end{subfigure}
\begin{subfigure}{.36\textwidth}\centering
\begin{tikzpicture}[scale=.67]
\begin{axis}[width=7cm, height=3.5cm, ylabel=$\#$cycles, xlabel=birth, ymin=0, ymax=2.5, xmin=0, xmax=6.5, xtick={1,2,3,4,5,6}]
\addplot[thick,cyan!70!black,mark=*,mark options={fill=white},samples at={2,4}] {1};
\addplot[thick,cyan!70!black,mark=*,samples at={1,2}] {1};
\addplot[thick,cyan!70!black,mark=*,mark options={fill=white},samples at={4,5}] {2};
\addplot[thick,cyan!70!black,mark=*,samples at={4,4}] {2};
\addplot[thick,cyan!70!black,mark=*,mark options={fill=white},samples at={5,6}] {1};
\addplot[thick,cyan!70!black,mark=*,samples at={5,5}] {1};
\end{axis}
\end{tikzpicture}
\caption{Betti Curve} 
\end{subfigure}
\begin{subfigure}{.36\textwidth}\centering
\begin{tikzpicture}[scale=.67]
\begin{axis}[width=7cm, height=3.5cm, ylabel=$\#$cycles, xlabel=birth, ymin=0, ymax=2.5, xmin=0, xmax=6.5, xtick={1,2,3,4,5,6}]
\addplot[thick,cyan!70!black,samples at={1,1.5}] {\x-1};
\addplot[thick,cyan!70!black,samples at={1.5,2}] {-\x+2};
\addplot[thick,cyan!70!black,samples at={2,3.5}] {\x-2};
\addplot[thick,cyan!70!black,samples at={3.5,4.5}] {-\x+5};
\addplot[thick,cyan!70!black,samples at={4.5,5}] {\x-4};
\addplot[thick,cyan!70!black,samples at={5,6}] {-\x+6};
\end{axis}
\end{tikzpicture}
\caption{Persistence Landscape} 
\end{subfigure}
\caption{A persistence diagram (a), its Betti Curve (b) and Persistence Landscape (c).}
\label{landscapeAndBetti}
\end{figure}
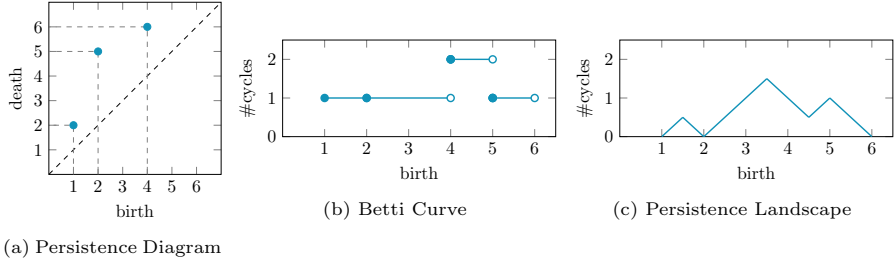

Let $D$ be a persistence diagram and $\alpha = (b_\alpha,d_\alpha)$ be a point in the diagram. Consider the following function: 
\[f_\alpha(x) \defeq \begin{cases} 1, &b_\alpha \leq x \leq d_\alpha \\ 0, &\text{otherwise} \end{cases}. \]
The \emph{Betti Curve} (Figure \ref{landscapeAndBetti}b) obtained from $D$ and $f_\alpha$ as the sum \[ Betti_D(x) \defeq \sum_{\alpha \in D} f_\alpha(x). \] 

Persistence landscape, introduced by Bubenik \cite{bubenik2015statistical}, is another statistical summary function that can be obtained from a persistence diagram $D$ (Figure \ref{landscapeAndBetti}c). Given $\alpha \in D$, let 
\[ g_\alpha(x) \defeq \begin{cases} x - b_\alpha, &\text{if } b_\alpha \leq x \leq (b_\alpha + d_\alpha)/2 \\ d_\alpha - x , &\text{if } (b_\alpha + d_\alpha)/2 < x \leq d_\alpha \\ 0, &\text{otherwise.} \end{cases} \]
Then, the $k$\textsuperscript{th} layer persistence landscape of $D$ is defined as \[ Landscape_D^k(x) \defeq \kmax_{\alpha \in D} g_\alpha(x) \] where $\kmax$ is the $k$\textsuperscript{th} maximum value among a finite set. In this study, we use the first layer persistence landscapes, and we will drop the superscript $k = 1$.

\section{Feature engineering on persistence diagrams}\label{sect:featureEngineering}

There are several ways to construct features from a persistence diagram, such as calculating the mean birth times of points in the diagram. However, these features are not stable in the sense that a slight perturbation in the dataset can create or remove several points with short lifetimes.

{In this study, we engineered featured based on topological data analysis methods using distances on persistence diagrams, persistent entropy, and Betti curves. We outline our method in this section. We also provide the pseudocodes for our feature engineering methods in Algorithm \ref{feature_engineering_algorithm}.}

\subsection{Via Bottleneck and Wasserstein distances}
Let $D_\emptyset$ be the diagram containing only the diagonal. For any diagram $D$, we can compute the Wasserstein distance for $p=1$: $W_1(D,D_\emptyset)$ and the bottleneck distance $W_\infty(D,D_\emptyset)$ as two features of a diagram. Let $l_x$ be the lifetime of a point $x \in D$. We can write them down explicitly since the perfect matching between any diagram $D$ with $D_\emptyset$ is obvious:
\begin{align*} 
W_1(D,D_\emptyset) = \frac{1}{\sqrt 2} \sum_{x \in D} \left(l_x \right) \qquad\qquad
W_\infty(D,D_\emptyset) = \frac{1}{\sqrt 2} \sup_{x \in D} \left(l_x \right)
\end{align*}

\subsection{Via persistent entropy}
Persistent Entropy is another summary statistic that can be derived from a persistence diagram \cite{atienza2019persistent}. A key feature of persistent entropy is that it is scale-invariant. The persistent entropy $PE(D)$ of a persistence diagram $D$ is given by \[ PE(D) \defeq \sum_{x \in D} - \frac{l_x}{L_D} \ln \left(\frac{l_x}{L_D}\right) \] where $L_D \defeq \sum_{x \in D} l_x$ is the sum of lifetimes. Whenever the diagram contained only the diagonal, the persistent entropy was assumed to be zero.

\subsection{Via norms}
Finally, we consider the Betti curve and first layer persistence landscape of a diagram $D$ as real-valued functions, and then compute their $L^1$ and $L^2$ norms. These are going to constitute our four new features constructed from these kernels.

\section{Sliding Windows and Delay Embeddings} \label{section_sliding_windows}

\subsection{Converting univariate series into multivariate series}

Let $x = (x_i: i=0,\ldots,N)$ be a univariate time series in $\mathbb R$. The sequence of \emph{sliding windows} on $x$ with a window shift of $k$ is a sequence of equally sized multivariate time series {(Algorithm \ref{subwindowing_algorithm})}:
\[ \left( (x_{kn}, x_{kn+1}, x_{kn+2}, \ldots, x_{kn+d-1}) :  n = 0,1,\ldots, \lfloor (N-d+1)/k \rfloor \right). \] 
Creating equally sized shorter windows from a large time series can be very useful for machine learning tasks. Note, however, that when the window shift is smaller than the window size, the sliding windows overlap. In such cases, to prevent data leakage between train and test sets, train-test splits should be carefully made.

\subsection{Delay embeddings}

The method of \emph{delay embeddings} allows us to embed a collection of time series of varying lengths into a euclidean space of a fixed dimension. {Under some assumptions, the topology of the delay embedding contains essential information about the nature of the time series. We refer the reader to \citet{perea2019topological} for theoretical justifications and several examples on the subject.} The $d$-dimensional delay embedding (also known as a \emph{state space reconstruction}) of $x$ with a shift of $k$ is the subset of $\mathbb{R}^d$ is given by {(Algorithm \ref{delay_embedding_algorithm})}
\[ \left\{(x_{kn}, x_{kn+1}, x_{kn+2}, \dots, x_{kn+d-1}) :  n = 0,1,\dots, \lfloor (N-d+1)/k \rfloor \right\}. \] The number $d$ is referred as the dimension or the size of the delay embedding.

One should note that if $y$ is a periodic signal, the time delay embedding on $y$ will follow a closed path on $\mathbb R^{d}$. For example, the time delay embeddings of $y = \{\cos x: x \in \mathbb T\}$, where $\mathbb T$ is a set of equally spaced points on $\mathbb R$, lies on a circle if the embedding dimension resonates with the frequency of $\cos x$ \cite{perea2015sliding}.

\subsubsection{An example}

\begin{figure}
\centering
\input{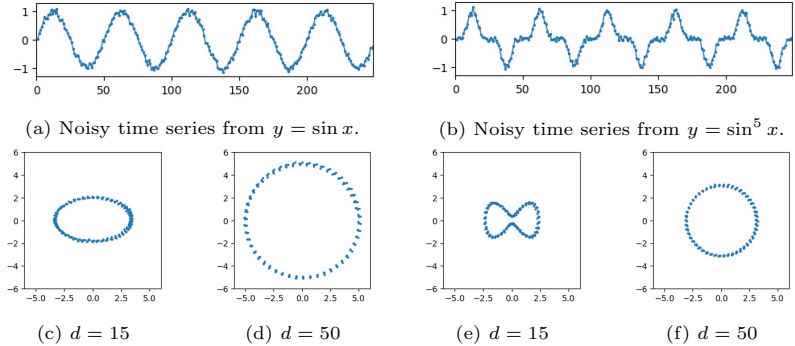}
\caption{Two noisy time series data along with their time delay embeddings for $d=15$ and $d=50$ under PCA visualization.}
\label{timeseries}
\end{figure}

The two time series of length $250$ in Figure \ref{timeseries} are sampled from five periods of $y=\sin x$ and $y=\sin^5x$ with additional noise, respectively. The time delay embeddings with $d=50$ for both time series are both circles with different radii. However, when the embedding dimension is small ($d=15$), the former is an ellipse whereas the latter is the boundary of an eyeglasses-shaped object. This idea shows us that we can distinguish the two time series for different embedding dimensions using persistent homology. 

\subsection{The subwindowing method}

\begin{figure}
\resizebox{\textwidth}{!}{\begin{tikzpicture}[yscale = .5]
\centering
\filldraw[gray!70!white] (0,1) rectangle node{\textcolor{black}{\footnotesize Full Signal}} + (15.5,.8);
\filldraw[gray!70!white] (0,0) rectangle node{\textcolor{black}{\footnotesize Window $1$}} + (6,.8);
\filldraw[gray!70!white] (1,-1) rectangle node{\textcolor{black}{\footnotesize Window $2$}} + (6,.8);
\filldraw[gray!70!white] (2,-2) rectangle node{\textcolor{black}{\footnotesize Window $3$}} + (6,.8);
\filldraw[gray!70!white] (0,-4) rectangle node{\textcolor{black}{\scalebox{.9}[1.0]{\footnotesize subwindow $1$}}} + (2,.8);
\filldraw[gray!70!white] (1,-5) rectangle node{\textcolor{black}{\scalebox{.9}[1.0]{\footnotesize subwindow $2$}}} + (2,.8);
\filldraw[gray!70!white] (2,-6) rectangle node{\textcolor{black}{\scalebox{.9}[1.0]{\footnotesize subwindow $3$}}} + (2,.8);
\filldraw[gray!70!white] (3,-7) rectangle node{\textcolor{black}{\scalebox{.9}[1.0]{\footnotesize subwindow $4$}}} + (2,.8);
\filldraw[gray!70!white] (4,-8) rectangle node{\textcolor{black}{\scalebox{.9}[1.0]{\footnotesize subwindow $5$}}} + (2,.8);
\filldraw[gray!70!white] (5,-9) rectangle node{\textcolor{black}{\scalebox{.9}[1.0]{\footnotesize subwindow $6$}}} + (2,.8);
\filldraw[gray!70!white] (6,-10) rectangle node{\textcolor{black}{\scalebox{.9}[1.0]{\footnotesize subwindow $7$}}} + (2,.8);

\draw [dashed] (0,-0.1) -- (0,-3.1);
\draw [dashed] (1,-1.1) -- (1,-3.1);
\draw [dashed] (2,-2.1) -- (2,-3.1);
\draw [dashed] (6,-2.1) -- (6,-7.1);
\draw [dashed] (7,-2.1) -- (7,-8.1);
\draw [dashed] (8,-2.1) -- (8,-9.1);

\draw [thick, ->] (8.1,-3.5) -- (8.5,-3.5) node{\footnotesize \hspace{1.5cm} features $1$};
\draw [thick, ->] (8.1,-4.5) -- (8.5,-4.5) node{\footnotesize \hspace{1.5cm} features $2$};
\draw [thick, ->] (8.1,-5.5) -- (8.5,-5.5) node{\footnotesize \hspace{1.5cm} features $3$};
\draw [thick, ->] (8.1,-6.5) -- (8.5,-6.5) node{\footnotesize \hspace{1.5cm} features $4$};
\draw [thick, ->] (8.1,-7.5) -- (8.5,-7.5) node{\footnotesize \hspace{1.5cm} features $5$};
\draw [thick, ->] (8.1,-8.5) -- (8.5,-8.5) node{\footnotesize \hspace{1.5cm} features $6$};
\draw [thick, ->] (8.1,-9.5) -- (8.5,-9.5) node{\footnotesize \hspace{1.5cm} features $7$};

\draw [densely dashed,->, shorten <= 2mm, shorten >= 6mm] (10.1,-3.5) -- (13, -4.5);
\draw [densely dashed,->, shorten <= 2mm, shorten >= 6mm] (10.1,-4.5) -- (13, -4.5);
\draw [densely dashed,->, shorten <= 2mm, shorten >= 6mm] (10.1,-5.5) -- (13, -4.5);
\draw [densely dashed,->, shorten <= 2mm, shorten >= 6mm] (10.1,-6.5) -- (13, -4.5);
\draw [densely dashed,->, shorten <= 2mm, shorten >= 6mm] (10.1,-7.5) -- (13, -4.5) node{\footnotesize \hspace{1.8cm} Window $1$ features};

\draw [densely dashed,->, shorten <= 2mm, shorten >= 6mm] (10.1,-4.5) -- (13, -6.5);
\draw [densely dashed,->, shorten <= 2mm, shorten >= 6mm] (10.1,-5.5) -- (13, -6.5);
\draw [densely dashed,->, shorten <= 2mm, shorten >= 6mm] (10.1,-6.5) -- (13, -6.5);
\draw [densely dashed,->, shorten <= 2mm, shorten >= 6mm] (10.1,-7.5) -- (13, -6.5);
\draw [densely dashed,->, shorten <= 2mm, shorten >= 6mm] (10.1,-8.5) -- (13, -6.5) node{\footnotesize \hspace{1.8cm} Window $2$ features};

\draw [densely dashed,->, shorten <= 2mm, shorten >= 6mm] (10.1,-5.5) -- (13, -8.5);
\draw [densely dashed,->, shorten <= 2mm, shorten >= 6mm] (10.1,-6.5) -- (13, -8.5);
\draw [densely dashed,->, shorten <= 2mm, shorten >= 6mm] (10.1,-7.5) -- (13, -8.5);
\draw [densely dashed,->, shorten <= 2mm, shorten >= 6mm] (10.1,-8.5) -- (13, -8.5);
\draw [densely dashed,->, shorten <= 2mm, shorten >= 6mm] (10.1,-9.5) -- (13, -8.5) node{\footnotesize \hspace{1.8cm} Window $3$ features};
\end{tikzpicture}}
\caption{The subwindowing method for feature construction}
\label{subwindowing_method}
\end{figure}
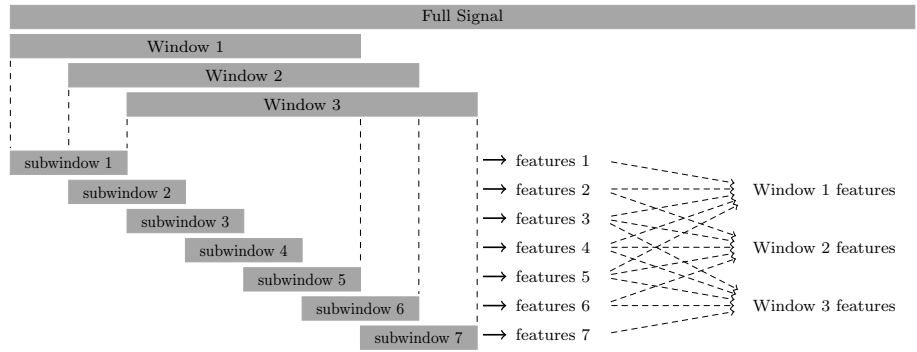

The subwindowing approach we developed in this article is essential to our analyses since it reduces noise and the required computational power. Assume that we would like to compute features on a set of sliding windows with a fixed window size and fixed window shift. Instead of directly computing features on each window, we can create sliding \emph{subwindows} on each window (Figure \ref{subwindowing_method}), then compute the features at the subwindows. Looking at the average and the standard deviation of the subwindow features, one can understand how the window behaves locally, and how this local behavior varies over time.

This approach is especially useful if we want to compute topological features on windows using persistent homology on delay embeddings. If a window contains a noisy part (as most physiological signals do), the corresponding delay embedding and the resulting persistence diagrams will be noisy as well. With subwindowing, the noise is trapped into a few subwindows, and its effects on the feature vector of a window diminish largely by computing the mean and standard deviation of feature vectors of subwindows.

Moreover, the subwindowing method has advantages for computation. Assume that the window features are obtained by finding the mean and standard deviation of the features from subwindows. Assume also that the subwindow shift is the same as the window shift as in Figure \ref{subwindowing_method}. Now, let Window $N$ and Window $N+1$ be two consecutive windows with subwindows $sw_1,\dots,sw_M$ and $sw_2,\dots,{sw_{M+1}}$, respectively. Let $f_i$ be the feature coming from $sw_i$. Then, the mean feature for Window $N$ is \[ \mu_N \defeq \frac{1}{M} \sum_{i=1}^M f_i \]
and the standard deviation of features for Window $N$ is \[\sigma_N \defeq \sqrt{\frac{1}{M} \sum_{i=1}^M (f_i - \mu_N)^2}  {= \sqrt{\frac{\sum_{i=1}^M f_i^2}{M} - \mu_N^2}.} \]
Observe that \[\mu_{N+1} = \frac{1}{M} \sum_{i=2}^{M+1} f_i = \left(\frac{1}{M} \sum_{i=1}^M f_i \right) + \frac{f_{M+1}-f_1}{M} = \mu_N + \frac{f_{M+1}-f_1}{M}. \]
Also,
\begin{align*}
\sigma^2_{N+1} = {\frac{\sum_{i=2}^{M+1} f_i^2}{M} - \mu_{N+1}^2 = \sigma^2_N + \frac{f_{M+1}^2 - f_1^2}{M} + \mu_N^2 - \mu_{N+1}^2}.
\end{align*}
So, $\mu_{N+1}$ can be written in terms of $\mu_N, f_1, f_{M+1}$ and $M$. Similarly, $\sigma_{N+1}$ can be written as a function of $\sigma_N,\mu_N,\mu_{N+1}, f_1, f_{M+1}$ and $M$ {(Algorithm \ref{rolling_mean_algorithm})}.

This means once the mean and standard deviation for a window is computed, these features for the next sliding windows can be computed continuously regardless of the window size. If the sampling frequency of the signal is high, and if the window size is large, then computing features directly from the windows can be difficult. Subwindowing allows us to compute the features at the same time regardless of the window size.

\section{Dataset Description} \label{sect:dataset_description}

In this study, we demonstrate the use of persistent homology for affect and stress recognition. {For this purpose, we used one synthetic and two publicly available real datasets. In the synthetic dataset, we aimed to simulate physiological signals from stress and non-stress conditions.} The real datasets, WESAD \cite{schmidt2018introducing} and DriveDB\footnote{Available at https://physionet.org/content/drivedb/1.0.0/ \cite{goldberger2000physiobank}} \cite{healey2005detecting} contain physiological signals from participants who were subjected to some stress and non-stress conditions in different environments. The datasets are shortly described below.

\subsection{Synthetic dataset}

The methodology that we used to generate our synthetic datasets is adapted from the methodologies used by \citet{umeda2017time}. In this dataset, we simulated physiological signals using Python's NeuroKit2 library \cite{Makowski2021neurokit} consisting of respiration (RESP) and electrocardiogram (ECG) signals. We generated samples with 120s of non-stress and 120s of stress condition for $20$ simulated participants. The signals were generated at 50 Hz. We use the hypothesis that a sustained elevated respiration rate, a sustained elevated heart rate, or an increased heart rate variability are indicators for stress condition.

In the first experiment, we simulated RESP signals with different respiratory rates for the baseline (non-stress) and stress conditions. The respiratory rate for the baseline was set to 15 respirations per minute (rpm), and we used different integer values from 16 rpm to 20 rpm for the stress condition. In the second experiment, the baseline condition contained ECG signals with different heart rates. The baseline heart rate was set to be 70 bpm, and the stress heart rate (HR) was tested for different integer values from 71 bpm to 75 bpm. The standard deviation of the HR parameter was fixed to be 1 for both conditions. The last experiment was similar to the second one except that this time we manipulated the standard deviation of the HR while keeping the HR constant. The baseline again had an HR of 70 bpm with a standard deviation set to 1, but this time while the stress condition also had an HR of 70 bpm, we used an increasing sequence of standard deviations of $2,3,4$, and $5$ in our experiments. In order to explore how the results are affected by noise, we repeated our experiments with two different noise parameters $0.1$ and $0.3$ defined by the NeuroKit2 library.

\subsection{The WESAD dataset}

Physiological recordings from a wrist (Empatica E4) and a chest (RespiBAN) worn device were collected from 15 participants during three different conditions: baseline, stress and amusement (Figure \ref{protocol}a). All participants started the experiment with the baseline condition in which they did normal activities such as reading a magazine or sitting at a table approximately for 20 minutes. In the stress condition, participants went under the Trier Social Stress Test (TSST), which included doing a 5-minute presentation to an audience and counting backwards from 2023 with steps of 17. The stress condition lasted about 10 minutes. The amusement condition consisted of watching a sequence of funny video clips, which lasted about 7 minutes. Each of the three conditions was followed by a meditation phase aiming to bring subjects to a neutral state. The order of stress and amusement conditions was counterbalanced across participants.

Acceleration (ACC), RESP, ECG, electrodermal activity (EDA), electromyography (EMG), and temperature (TEMP) signals were collected from the chest-worn device at 700 Hz. The signals from the wrist-worn device were ACC, blood volume pulse (BVP), EDA, and TEMP, with sampling frequencies 32 Hz, 64 Hz, 4 Hz, and 4 Hz, respectively.

\subsection{The DriveDB dataset}

In this study, a total of $17$ participants were recorded under three stress levels. The low and high stress conditions correspond to driving on the highway and in the city (Figure \ref{protocol}b). The experiment started with resting in the car,  then driving on the highway and in the city several times, and finally ending the experiment with the rest condition again. The experiment lasted about 60-90 minutes, depending on participants' driving speeds. The physiological signals that were recorded during the experiments ECG, EMG, galvanic skin response (GSR) from foot and hand, HR, and RESP. The sampling frequency for all signals was 15.5 Hz. Recordings of only 9 participants were analyzed since the markers that show the transition between different stress conditions were not available or legible for others.

\definecolor{baseline}{rgb}{0.54, 0.81, 0.94} 
\definecolor{stress}{rgb}{0.8, 0.0, 0.0} 
\definecolor{rest}{rgb}{0.91, 1.0, 1.0} 
\definecolor{meditation}{rgb}{0.56, 0.74, 0.56} 
\definecolor{amusement}{rgb}{0.93, 0.53, 0.18} 
\definecolor{highway}{rgb}{1.0, 0.75, 0.0} 

\begin{figure}
\centering
\begin{subfigure}{\textwidth}
\centering
\scalebox{.8}{
\begin{tikzpicture}[every node/.append style={shading=axis, rectangle, anchor=west, minimum width=1.5cm, shading angle=135, minimum height=1cm, draw=black}]
\node[left color=baseline, right color=baseline!30!white] (baseline) {\sffamily Baseline};
\node[left color=stress, right color=stress!30!white] (stress) at (baseline.east) {\sffamily Stress};
\node[left color=rest, right color=rest!30!white] (rest) at (stress.east) {\sffamily Rest};
\node[left color=meditation, right color=meditation!30!white] (meditation1) at (rest.east) {\sffamily Meditation I};
\node[left color=amusement, right color=amusement!30!white] (amusement) at (meditation1.east) {\sffamily Amusement};
\node[left color=meditation, right color=meditation!30!white] (meditation2) at (amusement.east) {\sffamily Meditation II};
\end{tikzpicture}}

\vspace{.1cm}

\scalebox{.8}{
\begin{tikzpicture}[every node/.append style={shading=axis, rectangle, anchor=west, minimum width=1.5cm, shading angle=135, minimum height=1cm, draw=black}]
\node[left color=baseline, right color=baseline!30!white] (baseline) {\sffamily Baseline};
\node[left color=amusement, right color=amusement!30!white] (amusement) at (baseline.east) {\sffamily Amusement};
\node[left color=meditation, right color=meditation!30!white] (meditation1) at (amusement.east) {\sffamily Meditation I};
\node[left color=stress, right color=stress!30!white] (stress) at (meditation1.east) {\sffamily Stress};
\node[left color=rest, right color=rest!30!white] (rest) at (stress.east) {\sffamily Rest};
\node[left color=meditation, right color=meditation!30!white] (meditation2) at (rest.east) {\sffamily Meditation II};
\end{tikzpicture}} 
\caption{} \end{subfigure}

\vspace{.3cm}

\begin{subfigure}{\textwidth}
\centering
\scalebox{.8}{
\begin{tikzpicture}[every node/.append style={shading=axis, rectangle, anchor=west, minimum width=1.5cm, shading angle=135, minimum height=1cm, draw=black}]
\node[left color=baseline, right color=baseline!30!white] (relax1) {\sffamily Rest};
\node[left color=stress, right color=stress!30!white] (city1) at (relax1.east) {\sffamily City 1};
\node[left color=highway, right color=highway!30!white] (highway1) at (city1.east) {\sffamily Highway 1};
\node[left color=stress, right color=stress!30!white] (city2) at (highway1.east) {\sffamily  City 2};
\node[left color=highway, right color=highway!30!white] (highway2) at (city2.east) {\sffamily Highway 2};
\node[left color=stress, right color=stress!30!white] (city3) at (highway2.east) {\sffamily City 3};
\node[left color=baseline, right color=baseline!30!white] (relax2) at (city3.east) {\sffamily Rest};
\end{tikzpicture}}
\caption{} \end{subfigure}
\caption{The study protocols for WESAD (a) and DriveDB (b) datasets.}
\label{protocol}
\end{figure}
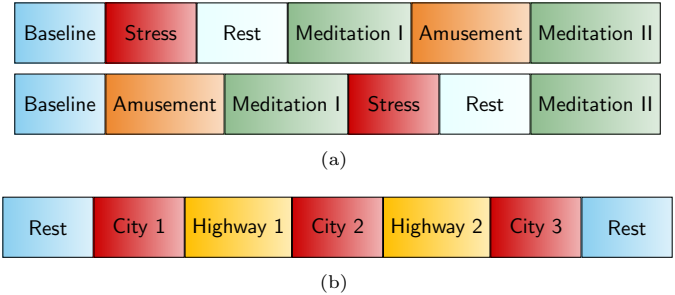

\section{Methodology}\label{sect:methodology}

All signals from all datasets were split by stress condition and by participant. To reduce computational time, all signals with a sampling rate higher than 100 Hz (\emph{i.e.} chest signals from WESAD) were downsampled to 100 Hz. Moreover, signals from DriveDB dataset were upsampled from 15.5 Hz to 16 Hz to ensure consistency in our method. Resampling from 700 Hz to 100 Hz was done by selecting every 7th element in the discrete time series. Resampling from 15.5 Hz to 16 Hz was done by upsampling to 496 Hz using linear interpolation, then downsampling by selecting every 31st element. The ACC data in WESAD contained three time series for the accelerations in $x,y,$ and $z$ axes; we averaged them to get a single time series.

\subsection{Sliding windows and subwindows}

Our goal is to create topological features from the windows and use them for affect and stress recognition. Sliding windows with duration of $60$ seconds and a shift of $2$ seconds were created for all datasets. {For our initial analyses, we specifically selected a window size of $60$ seconds to make our findings comparable with the original WESAD study. We also varied the window size to test how accuracy is affected. Longer windows produce higher accuracies (Table \ref{subwindow_size}). However, in real applications, one might want to keep the time durations short for stress detection.}

\begin{table}[H]
\centering
\begin{tabular}{@{}lllllll@{}}
\toprule
                                &   & \multicolumn{5}{c}{Window size}       \\ 
                                                                &   & 10    & 20    & 30    & 60    & 120   \\ \cmidrule{3-7} 
\multirow{3}{*}{\begin{tabular}[c]{@{}c@{}}Subwindow \\ size\end{tabular}} & 3 & 74.38 & 77.92 & 79.08 & 80.67 & 85.18 \\
                                & 4 & 75.17 & 76.63 & 78.36 & 81.35 & 85.01 \\
                                & 5 & 75.39 & 77.95 & 80.07 & 81.25 & 84.29 \\ \bottomrule
\end{tabular}
\caption{Classification accuracies across different subwindow and window sizes for WESAD.}
\label{subwindow_size}
\end{table}

In each window, subwindows with 4s duration and 2s shift were created. {This choice of subwindow size was the same regardless of the physiological signal in order to keep the algorithm as simple as possible. To obtain consistent and reliable persistence diagrams, the subwindow size should be large enough to exhibit the periodicity in the delay embedding (Figure \ref{subwindow_persistence_diagrams_pipeline}). Our empirical tests showed that the subwindow size of 4 seconds is long enough to capture the local information about the physiological signal including periodicity, yet it is also short enough to make the computation of persistent homology coming from the delay embeddings feasible. In Table \ref{subwindow_size}, we show the changes in accuracies across different window and subwindow sizes using an SVM classifier. We give the feature engineering and cross-validation methods that we used to obtain the accuracies in Table \ref{subwindow_size} later in this section. The choice of subwindow shift is more of a computational issue: if the shift drops from 2s to 1s, the number of subwindows, hence the time required for persistent homology computations are doubled.}

\begin{figure}[H]
\centering
\includegraphics[width = .8\textwidth]{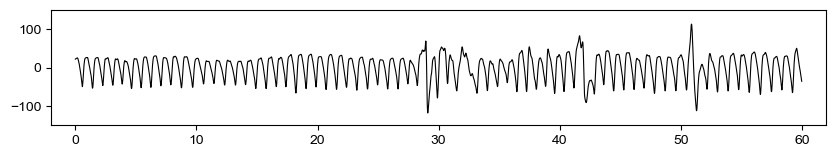}
\caption{A sample 60-second BVP signal from the baseline condition of WESAD dataset.}
\label{noisyBVP}
\end{figure}

The subwindowing method was useful in our study for several reasons. First, stressful events usually induce irregular physiological responses. For example, a typical response to stress is high heart rate variability. So, looking at how the subwindows behave across a window is informative for the current task. That is, subwindowing helps us understand the local topology of the window. Secondly, even in non-stress conditions, brief yet powerful noises are common due to participants' coughing, sudden movements, etc. (e.g. Figure \ref{noisyBVP}). Third, the delay embedding of a window is a very large dataset (especially if the sampling frequency is high such as 100 Hz), making the computation of persistent homology impractical.

\subsection{Delay embeddings and persistent homology of subwindows}

\begin{figure}[H]
  \centering
  \resizebox{\textwidth}{!}{\input{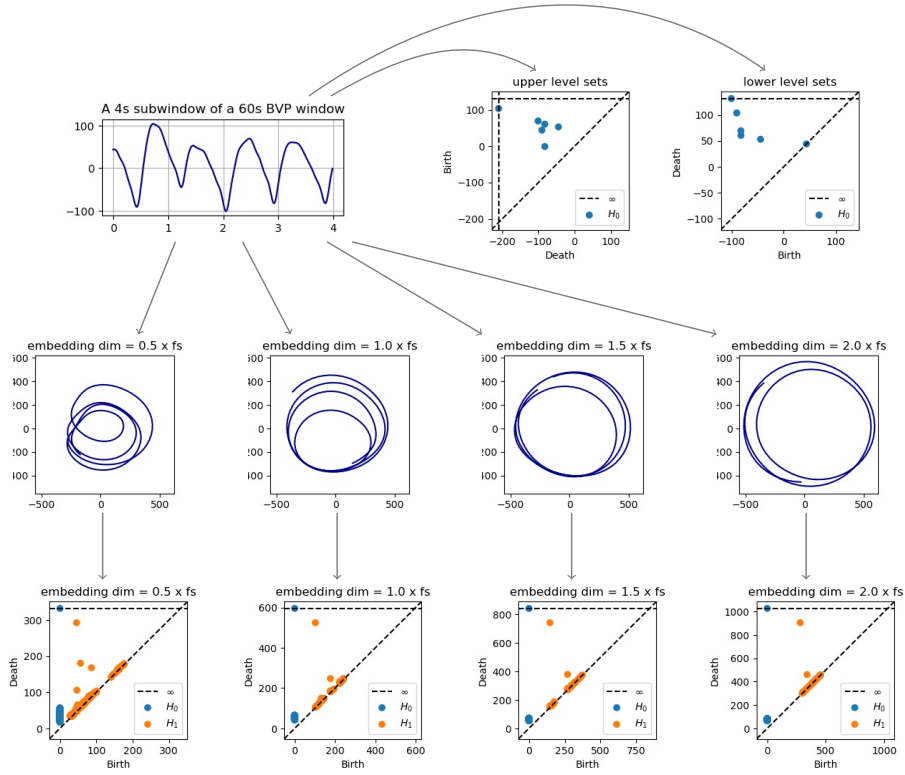}}
  \caption{The pipeline for the computation of persistence diagrams from a subwindow.}
  \label{subwindow_persistence_diagrams_pipeline}
\end{figure}

As we noted earlier in Section \ref{section_sliding_windows}, different embedding dimensions detect different topological information about the time series (Figure \ref{timeseries}). We used $4$ levels of delay embedding dimensions (see Figure \ref{subwindow_persistence_diagrams_pipeline}): .5\emph{fs}, \emph{fs}, 1.5\emph{fs}, 2\emph{fs}  where \emph{fs} is the sampling frequency of the particular signal. For example, for a signal with \emph{fs} $ = 100$Hz,  the set of embedding dimensions were $\{ 50,100,150,200\}$.

After converting each subwindow to $4$ different point clouds, the persistence diagrams of the induced Rips filtrations were computed for a maximum {homology} dimension of $1$ {(Algorithm \ref{persistent_homology_algorithm})}. Higher dimensional persistence diagrams were not computed since they require much more computational power, and we wanted to restrict our attention to connected components and one dimensional holes of the delay embeddings, but not to higher dimensional topological features.

In addition to the diagrams formed by the delay embeddings, two more persistence diagrams were computed: the 0 dimensional persistence diagrams created by the upper and lower level sets of the subwindows. Note that higher dimensional persistent homology cannot be computed from the subwindows because the subwindows are univariate. Persistent homology of delay embeddings and level sets were computed using the Ripser \cite{tralie2018ripser} and Dionysus-2 libraries \cite{dionysus} of the Python programming language \cite{van1995python}, respectively.

\subsection{Feature engineering}

Our methods have so far provided $6$ persistence diagrams, $4$ from delay embeddings and $2$ for upper and lower level sets, for each subwindow.  Then using the methods we outlined in Section~\ref{sect:featureEngineering}, a total of $7$ features were created from one homology class in a single persistence diagram. Since each subwindow yielded a total of $6$ persistence diagrams and $4$ diagrams coming from delay embeddings contained two homology classes, the total number of features created using persistent homology was $70$.

In order to compute features for the 60s windows, the mean and standard deviation of the features obtained from (4s) subwindows were calculated. This allowed us to know how the signal behaves locally, and how this local behavior varies over a longer period. We also did the same for different window sizes (10, 20, 30, 60, 120, 180, 240, and 300 seconds), and specifically looked at how the recognition accuracies change accordingly.

\begin{figure}[H]
\centering
\includegraphics[width = .4\textwidth]{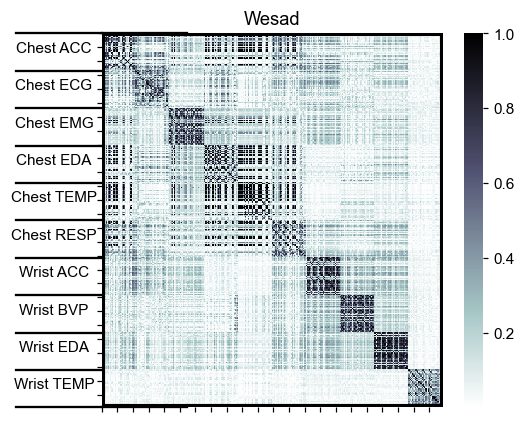} \hspace{1cm}
\includegraphics[width = .4\textwidth]{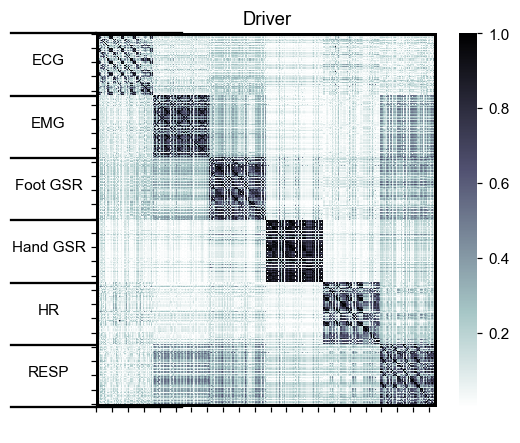}
\caption{Correlation heatmaps for WESAD and DriveDB.}
\label{correlations}
\end{figure}

The learning algorithms were tested on several subsets of features. For each of the four delay embedding sizes ($.5$\emph{fs}, $1$\emph{fs}, $1.5$\emph{fs}, $2$\emph{fs}), features from homology dimension zero (H0) and one (H1) were trained individually and together. Similarly, features coming from upper and lower level sets were trained one by one and together. Then features from all delay embeddings and level sets were combined and used as a full feature set. Before training the algorithms, constant features (\emph{e.g.} full zeros) and features with a correlation higher than $.9$ were removed. {This step was specifically essential since some machine learning algorithms such as Linear Discriminant Analysis are very sensitive to multicollinearity. The correlation heatmaps show that features were moderately correlated within sensors, and weakly correlated or not correlated between sensors (Figure \ref{correlations}).} Then, features were normalized to the range $[0,1]$ on both the training and the test set. 

\subsection{Learning algorithms}

The original WESAD study used five classifiers: Decision tree, Random forest (RF), AdaBoost Decision Tree (AB), Linear discriminant analysis (LDA), and $k$-nearest neighbor. Three of them (RF, AB, LDA) attained the highest accuracies for some signals. In addition to these three, we used a support vector classifier (SVC).

The tree-based models (RF and AB) were trained on 100 trees with a maximum depth of 5, and the SVC model was trained with a linear kernel and regularization parameter set to $0.1$. We used the scikit-learn library \cite{scikit-learn} implementations of the learning algorithms. For the reproducibility of our findings, we set the parameter \texttt{random\_state} to 0 in our stochastic models.

\subsection{Cross-validation}

\begin{figure}[H]
 \centering
\centering
\scalebox{.7}{
\begin{tikzpicture}[scale = .85]
\draw [fill=orange] (0,5) rectangle (2,5.7) node[pos=.5] {Subject 1};
\draw [fill=gray!50!white] (2,5) rectangle (4,5.7) node[pos=.5] {Subject 2};
\draw [fill=gray!50!white] (4,5) rectangle (6,5.7) node[pos=.5] {Subject 3};
\draw [fill=gray!50!white] (6,5) rectangle (8,5.7) node[pos=.5] {$\cdots$};
\draw [fill=gray!50!white] (8,5) rectangle (10,5.7) node[pos=.5] {Subject $N$};

\draw [fill=gray!50!white] (0,4) rectangle (2,4.7) node[pos=.5] {Subject 1};
\draw [fill=orange] (2,4) rectangle (4,4.7) node[pos=.5] {Subject 2};
\draw [fill=gray!50!white] (4,4) rectangle (6,4.7) node[pos=.5] {Subject 3};
\draw [fill=gray!50!white] (6,4) rectangle (8,4.7) node[pos=.5] {$\cdots$};
\draw [fill=gray!50!white] (8,4) rectangle (10,4.7) node[pos=.5] {Subject $N$};

\draw [fill=gray!50!white] (0,3) rectangle (2,3.7) node[pos=.5] {Subject 1};
\draw [fill=gray!50!white] (2,3) rectangle (4,3.7) node[pos=.5] {Subject 2};
\draw [fill=orange] (4,3) rectangle (6,3.7) node[pos=.5] {Subject 3};
\draw [fill=gray!50!white] (6,3) rectangle (8,3.7) node[pos=.5] {$\cdots$};
\draw [fill=gray!50!white] (8,3) rectangle (10,3.7) node[pos=.5] {Subject $N$};

\node[below] at (5,2.9) {$\cdots$};

\draw [fill=gray!50!white] (0,1.5) rectangle (2,2.2) node[pos=.5] {Subject 1};
\draw [fill=gray!50!white] (2,1.5) rectangle (4,2.2) node[pos=.5] {Subject 2};
\draw [fill=gray!50!white] (4,1.5) rectangle (6,2.2) node[pos=.5] {Subject 3};
\draw [fill=gray!50!white] (6,1.5) rectangle (8,2.2) node[pos=.5] {$\cdots$};
\draw [fill=orange] (8,1.5) rectangle (10,2.2) node[pos=.5] {Subject $N$};

\draw [fill=gray!50!white] (3,.9) rectangle (3.3,1.2) node[right, yshift = -1mm] {Train};
\draw [fill=orange] (5,.9) rectangle (5.3,1.2) node[right, yshift = -1mm] {Test};

\draw[thick,->,shorten >=2ex] (10.1,5.4) -- (12, 3.4);
\draw[thick,->,shorten >=2ex] (10.1,4.4) -- (12, 3.4);
\draw[thick,->,shorten >=2ex] (10.1,3.4) -- (12, 3.4);
\draw[thick,->,shorten >=2ex] (10.1,1.9) -- (12, 3.4);

\node[right, ,text width=1.2cm] at (12,3.4) {Averaged};
\end{tikzpicture}
}
  \caption{Leave One Subject Out Cross Validation (LOSOCV).}
  \label{losocv}
\end{figure}
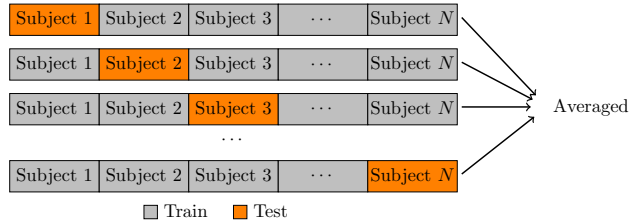

For all datasets, we used binary (stress vs. non-stress) classification models, but for WESAD and DriveDB  datasets, we also used ternary classification models. Leave One Subject Out Cross-Validation (LOSOCV) method was used for all datasets. This method is similar to the k-fold cross validation if we let $k$ be the number of participants, and each fold to be a participant's data. That is, the learning algorithm is trained on all subjects but one, tested on the remaining subject, and then the results are averaged (Figure \ref{losocv}). The biggest conceptual advantage of this method is that it helps us know how the model performs on a previously unseen participant. Furthermore, no data leakage between the train and test sets happens even when the sliding windows overlap highly.

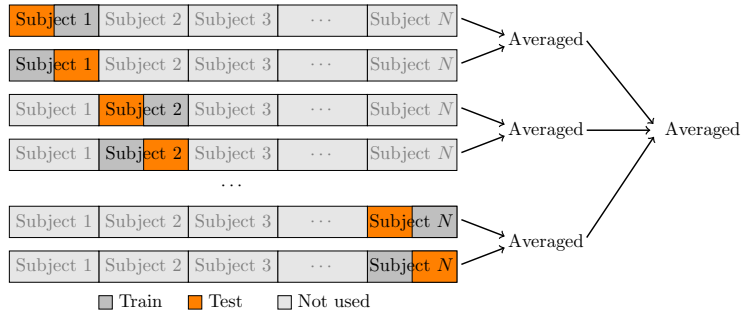
\begin{figure}[H]
 \centering
\centering
\scalebox{.7}{
\begin{tikzpicture}[scale=.85]

\draw [fill=orange] (0,5) rectangle (2,5.7) node[pos=.5] {};
\draw [fill=gray!50!white] (1,5) rectangle (2,5.7) node[pos=.5] {};
\node[right] at (0,5.35) {Subject 1};
\draw [fill=gray!20!white] (2,5) rectangle (4,5.7) node[pos=.5] {\textcolor{gray}{Subject 2}};
\draw [fill=gray!20!white] (4,5) rectangle (6,5.7) node[pos=.5] {\textcolor{gray}{Subject 3}};
\draw [fill=gray!20!white] (6,5) rectangle (8,5.7) node[pos=.5] {\textcolor{gray}{$\cdots$}};
\draw [fill=gray!20!white] (8,5) rectangle (10,5.7) node[pos=.5] {\textcolor{gray}{Subject $N$}};

\draw [fill=gray!50!white] (0,4) rectangle (2,4.7) node[pos=.5] {};
\draw [fill=orange] (1,4) rectangle (2,4.7) node[pos=.5] {};
\node[right] at (0,4.35) {Subject 1};
\draw [fill=gray!20!white] (2,4) rectangle (4,4.7) node[pos=.5] {\textcolor{gray}{Subject 2}};
\draw [fill=gray!20!white] (4,4) rectangle (6,4.7) node[pos=.5] {\textcolor{gray}{Subject 3}};
\draw [fill=gray!20!white] (6,4) rectangle (8,4.7) node[pos=.5] {\textcolor{gray}{$\cdots$}};
\draw [fill=gray!20!white] (8,4) rectangle (10,4.7) node[pos=.5] {\textcolor{gray}{Subject $N$}};

\draw[thick,->,shorten >=2ex] (10.1,5.4) -- (11.4, 4.9);
\draw[thick,->,shorten >=2ex] (10.1,4.4) -- (11.4, 4.9);
\node[right, ,text width=1.2cm] at (11,4.9) {Averaged};

\draw [fill=orange] (2,3) rectangle (4,3.7) node[pos=.5] {};
\draw [fill=gray!50!white] (3,3) rectangle (4,3.7) node[pos=.5] {};
\node[right] at (2,3.35) {Subject 2};
\draw [fill=gray!20!white] (0,3) rectangle (2,3.7) node[pos=.5] {\textcolor{gray}{Subject 1}};
\draw [fill=gray!20!white] (4,3) rectangle (6,3.7) node[pos=.5] {\textcolor{gray}{Subject 3}};
\draw [fill=gray!20!white] (6,3) rectangle (8,3.7) node[pos=.5] {\textcolor{gray}{$\cdots$}};
\draw [fill=gray!20!white] (8,3) rectangle (10,3.7) node[pos=.5] {\textcolor{gray}{Subject $N$}};

\draw [fill=gray!50!white] (2,2) rectangle (4,2.7) node[pos=.5] {};
\draw [fill=orange] (3,2) rectangle (4,2.7) node[pos=.5] {};
\node[right] at (2,2.35) {Subject 2};
\draw [fill=gray!20!white] (0,2) rectangle (2,2.7) node[pos=.5] {\textcolor{gray}{Subject 1}};
\draw [fill=gray!20!white] (4,2) rectangle (6,2.7) node[pos=.5] {\textcolor{gray}{Subject 3}};
\draw [fill=gray!20!white] (6,2) rectangle (8,2.7) node[pos=.5] {\textcolor{gray}{$\cdots$}};
\draw [fill=gray!20!white] (8,2) rectangle (10,2.7) node[pos=.5] {\textcolor{gray}{Subject $N$}};

\draw[thick,->,shorten >=2ex] (10.1,3.4) -- (11.4, 2.9);
\draw[thick,->,shorten >=2ex] (10.1,2.4) -- (11.4, 2.9);
\node[right, ,text width=1.2cm] at (11,2.9) {Averaged};

\node[below] at (5,1.9) {$\cdots$};

\draw [fill=orange] (8,.5) rectangle (10,1.2) node[pos=.5] {};
\draw [fill=gray!50!white] (9,.5) rectangle (10,1.2) node[pos=.5] {};
\node[right] at (7.9,.85) {Subject $N$};
\draw [fill=gray!20!white] (2,.5) rectangle (4,1.2) node[pos=.5] {\textcolor{gray}{Subject 2}};
\draw [fill=gray!20!white] (4,.5) rectangle (6,1.2) node[pos=.5] {\textcolor{gray}{Subject 3}};
\draw [fill=gray!20!white] (6,.5) rectangle (8,1.2) node[pos=.5] {\textcolor{gray}{$\cdots$}};
\draw [fill=gray!20!white] (0,.5) rectangle (2,1.2) node[pos=.5] {\textcolor{gray}{Subject 1}};

\draw [fill=gray!50!white] (8,.5-1) rectangle (10,1.2-1) node[pos=.5] {};
\draw [fill=orange] (9,.5-1) rectangle (10,1.2-1) node[pos=.5] {};
\node[right] at (7.9,.85-1) {Subject $N$};
\draw [fill=gray!20!white] (2,.5-1) rectangle (4,1.2-1) node[pos=.5] {\textcolor{gray}{Subject 2}};
\draw [fill=gray!20!white] (4,.5-1) rectangle (6,1.2-1) node[pos=.5] {\textcolor{gray}{Subject 3}};
\draw [fill=gray!20!white] (6,.5-1) rectangle (8,1.2-1) node[pos=.5] {\textcolor{gray}{$\cdots$}};
\draw [fill=gray!20!white] (0,.5-1) rectangle (2,1.2-1) node[pos=.5] {\textcolor{gray}{Subject 1}};

\draw[thick,->,shorten >=2ex] (10.1,3.4-2.5) -- (11.4, 2.9-2.5);
\draw[thick,->,shorten >=2ex] (10.1,2.4-2.5) -- (11.4, 2.9-2.5);
\node[right, ,text width=1.2cm] at (11,2.9-2.5) {Averaged};

\draw[thick,->,shorten >=1ex] (12.9,4.9) -- (14.5, 2.9);
\draw[thick,->,shorten >=1ex] (12.9,2.9) -- (14.5, 2.9);
\draw[thick,->,shorten >=1ex] (12.9,.5) -- (14.5, 2.9);
\node[right, ,text width=1.2cm] at (14.5,2.9) {Averaged};

\draw [fill=gray!50!white] (2,.9-2) rectangle (2.3,1.2-2) node[right, yshift = -1mm] {Train};
\draw [fill=orange] (4,.9-2) rectangle (4.3,1.2-2) node[right, yshift = -1mm] {Test};
\draw [fill=gray!20!white] (6,.9-2) rectangle (6.3,1.2-2) node[right, yshift = -1mm] {Not used};
\end{tikzpicture}
}
  \caption{Intra-subject cross-validation.}
  \label{intracv}
\end{figure}

In order to assess how much of the performance is due to individual differences, we also used an \emph{intra-subject} cross-validation. For this, we consider only data from a single participant, split each condition in half, then use the first halves to predict the second, and \emph{vice versa} (Figure \ref{intracv}). The mean accuracy gives the accuracy for that subject, and averaging over all subjects gives the overall accuracy.

\section{{Results}} \label{sect:results}

\subsection{Synthetic dataset}

For the synthetic dataset, we used an SVM classifier on all topological features for every signal. We used 60s windows and LOSOCV method to make our results comparable across datasets.

\begin{figure}[H]
\centering
\includegraphics[scale=.5]{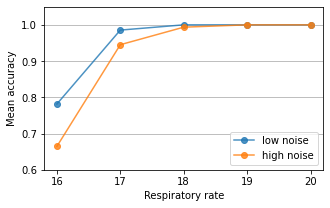}
\caption{Recognition accuracies as a function of respiratory rate with baseline 15 rpm.}
\label{simulated_rsp_change}
\end{figure}

For the RESP signal, our baseline respiratory rate was 15 respirations per minute (rpm). To simulate stress conditions, we increased the respiratory rates from 16 rpm to 20 rpm with increments of 1 rpm. The success rate of our models in distinguishing baseline from the stress increased as we increased the stress levels measured by an increase in the respiratory rate. Our models were highly successful for 16 rpm. Furthermore, our models worked almost perfectly for 17 rpm and higher even in the presence of high noise (Figure \ref{simulated_rsp_change}). 

\begin{figure}[H]
\centering
\includegraphics[scale=.5]{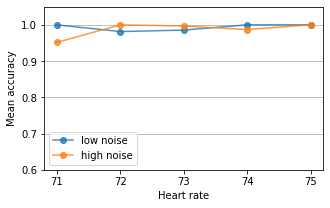}
\caption{Recognition accuracies as a function of heart rate with baseline 70 bpm.}
\label{simulated_hr_change}
\end{figure}

For the ECG signals, the baseline heart rate was 70 bpm. To simulate the stress conditions, we gradually increased the heart rate from 71 bpm to 75 bpm with increments of 1 bpm. Our models distinguished the baseline from all stress levels nearly perfectly, even in the presence of high noise (Figure~\ref{simulated_hr_change}). 

\begin{figure}[H]
\centering
\includegraphics[scale=.5]{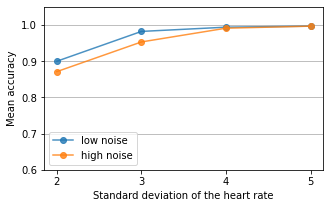}
\caption{Recognition accuracies as a function of standard deviation of the heart rate with baseline 1.}
\label{simulated_hrv_change}
\end{figure}

In our last analyses, we changed the underlying stress indicator.  For this part of the study, the variability in the heart rate is assumed to be the main indicator of a stress condition. For this synthetic dataset, the mean heart rate for the baseline was 70 bpm with a standard deviation of 1. To simulate the stress condition, we increased the standard deviation from 2 to 5 with increments of 1 as we kept the mean heart rate at 70 bpm. While our model was very successful in distinguishing the low stress condition (standard deviation 2), they performed nearly perfectly in high stress conditions (standard deviations 3 to 5) even in the presence of high noise (Figure~\ref{simulated_hrv_change}).

\subsection{WESAD dataset}

Our findings from LOSOCV showed that our automatically created topological features were as effective as the signal-specific features in distinguishing different affect state conditions. We show our results in Table \ref{Wesad_LOSOCV_ternary_table} and Table \ref{Wesad_LOSOCV_binary_table}.

For the ternary classification problem (Table \ref{Wesad_LOSOCV_ternary_table}), a support vector classifier on all topological features yielded 81.35\% accuracy and with an $F_1$-score of 73.44\%. For almost all signals in WESAD, our automatically created features identified the affective states better than the original study. The most dramatic difference was in the ACC signals. For the chest ACC, our model performed 17\% better, while for the wrist ACC, our model performed 11\% better. This was followed by ECG, EMG, and wrist EDA where our model performed 10\%, 6\%, and 10\% better than the original study. In almost all cases, our model performed the best when all topological features were combined as a full feature vector using an SVM classifier.

\begin{table}[H]
\centering
\resizebox{.8\textwidth}{!}{\begin{tabular}{@{}llclclcc@{}}
\toprule
 &
  \multicolumn{1}{c}{} &
  \multicolumn{2}{c}{\begin{tabular}[c]{@{}c@{}}Delay\\ Embeddings\end{tabular}} &
  \multicolumn{2}{c}{\begin{tabular}[c]{@{}c@{}}Level\\ Sets \end{tabular}} &
   &
   \\ \cmidrule(lr){3-4}
\cmidrule(lr){5-6} &
  \multicolumn{1}{c}{Clf} &
  Acc &
  \multicolumn{1}{c}{Dgm} &
  Acc &
  \multicolumn{1}{c}{Dgm} &
  \begin{tabular}[c]{@{}c@{}}All\\ Dgms\end{tabular} &
  \begin{tabular}[c]{@{}c@{}}Original\\ Findings\end{tabular} \\ \midrule
\textit{Chest}    & \multicolumn{1}{c}{} &                 & \multicolumn{1}{c}{} &       & \multicolumn{1}{c}{} &                 &                 \\
\hspace{2mm} ACC  & SVC                  & 72.34           & $2\fs$               & 69.77 & Upper                    & \highest{74.47} & 56.56           \\
\hspace{2mm} ECG  & SVC                  & \highest{76.27} & $1\fs$               & 66.11 & Upper                    & 72.72           & 66.29           \\
\hspace{2mm} EMG  & SVC                  & 59.00           & $1\fs$               & 54.64 & Both                    & \highest{59.67} & 53.99           \\
\hspace{2mm} EDA  & LDA                  & 66.31           & $.5\fs,H0$           & 66.56 & Upper                    & \highest{70.03} & 67.07           \\
\hspace{2mm} TEMP & LDA                  & 54.75           & $1\fs,H1$            & 53.02 & Upper                    & 48.92           & \highest{55.68} \\
\hspace{2mm} RESP & SVC                  & 68.46           & $2\fs$               & 70.73 & Both                    & \highest{75.57} & 72.37           \\ \midrule
\textit{Wrist}    &                      &                 &                      &       &                      &                 &                 \\
\hspace{2mm} ACC  & RF                   & 68.02           & $.5\fs$              & 67.28 & Upper                    & \highest{68.65} & 57.20           \\
\hspace{2mm} BVP  & SVC                  & 71.64           & $1\fs$               & 60.67 & Lower                    & \highest{73.41} & 70.17           \\
\hspace{2mm} EDA  & RF                   & 69.23           & $2\fs$               & 70.50 & Both                    & \highest{72.07} & 62.32           \\
\hspace{2mm} TEMP & LDA                  & 55.87           & .$5\fs$              & 54.79 & Upper                    & 54.93           & \highest{58.96} \\ \midrule
All chest         & SVC                  & \highest{78.85} & $1\fs,H0$            & 75.34 & Upper                    & 77.96           & 76.50           \\
All wrist         & RF                   & 73.93           & $2\fs$               & 71.94 & Upper                    & 74.72           & \highest{75.21} \\
All               & SVC                  & 80.63           & $1\fs$               & 80.56 & Lower                    & \highest{81.35} & 79.57           \\ \bottomrule
\end{tabular} 
}

\caption{Ternary classification problem accuracies for WESAD.}
\label{Wesad_LOSOCV_ternary_table}
\end{table}

\begin{figure}[H]
\centering
\includegraphics[height = 4cm]{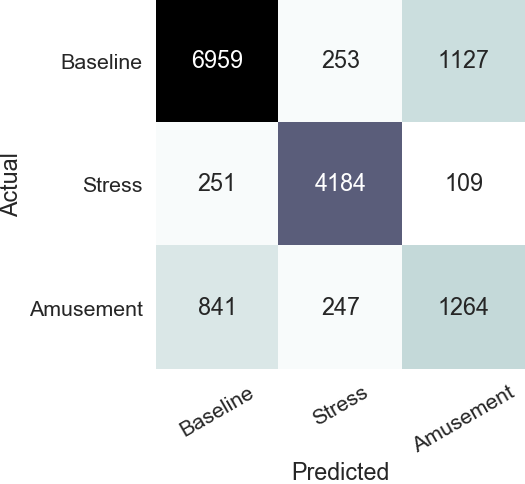}
\caption{Three-class problem confusion matrix for WESAD.}
\label{Wesad_confusion}
\end{figure}

We have got a clear separation in the confusion matrix (Figure \ref{Wesad_confusion}). The model performed better in distinguishing stress from non-stress. Most classification errors were between the baseline and amusement conditions.

For the binary classification task (Table \ref{Wesad_LOSOCV_binary_table}), the highest accuracy and the corresponding $F_1$-score were 94.46\% and 93.26\%. For nearly all physiological signals, we obtained higher accuracies with topological features. Again, the ACC signals captured the stress state very well: the accuracy for the chest ACC increased by 14\% and for the wrist ACC by 12\% compared to the original study. The improvements for the ECG, EMG, and wrist EDA signals were 3\%, 6\%,  and 5\% for the binary task.

\begin{table}[H]
\centering
\resizebox{.8\textwidth}{!}{\begin{tabular}{@{}llclclcc@{}}
\toprule
 &
  \multicolumn{1}{c}{} &
  \multicolumn{2}{c}{\begin{tabular}[c]{@{}c@{}}Delay\\ Embeddings\end{tabular}} &
  \multicolumn{2}{c}{\begin{tabular}[c]{@{}c@{}}Level\\ Sets\end{tabular}} &
   &
   \\ \cmidrule(lr){3-4}
\cmidrule(lr){5-6} &
  \multicolumn{1}{c}{Clf} &
  Acc &
  \multicolumn{1}{c}{Dgm} &
  Acc &
  \multicolumn{1}{c}{Dgm} &
  \begin{tabular}[c]{@{}c@{}}All\\ Dgms\end{tabular} &
  \begin{tabular}[c]{@{}c@{}}Original\\ Findings\end{tabular} \\ \midrule
\textit{Chest}    & \multicolumn{1}{c}{} &                 & \multicolumn{1}{c}{} &       & \multicolumn{1}{c}{} &                 &                 \\
\hspace{2mm} ACC  & RF                   & \highest{87.69} & $.5\fs,H0$           & 84.82 & Both                    & 84.18           & 73.87           \\
\hspace{2mm} ECG  & LDA                  & \highest{88.70} & $1\fs,H0$            & 81.62 & Both                    & 85.23           & 85.44           \\
\hspace{2mm} EMG  & LDA                  & \highest{73.07} & $1\fs$               & 68.03 & Lower                    & 69.75           & 67.10           \\
\hspace{2mm} EDA  & LDA                  & 76.57           & $1\fs,H1$            & 81.55 & Lower                    & 81.49           & \highest{81.70} \\
\hspace{2mm} TEMP & SVC                  & \highest{70.19} & $.5\fs$              & 70.19 & Both                    & 68.07           & 69.49           \\
\hspace{2mm} RESP & LDA                  & 81.72           & $.5\fs$              & 87.75 & Both                    & \highest{90.10} & 88.09           \\ \midrule
\textit{Wrist}    &                      &                 &                      &       &                      &                 &                 \\
\hspace{2mm} ACC  & RF                   & 82.95           & $.5\fs$              & 82.81 & Both                    & \highest{83.52} & 71.69           \\
\hspace{2mm} BVP  & LDA                  & 85.21           & $1\fs$               & 72.51 & Lower                    & 84.03           & \highest{85.83} \\
\hspace{2mm} EDA  & RF                   & 81.19           & $2\fs$               & 84.01 & Upper                    & \highest{85.11} & 79.71           \\
\hspace{2mm} TEMP & RF                   & \highest{71.44} & $.5\fs$              & 70.00 & Both                    & 70.02           & 69.24           \\ \midrule
All chest         & SVC                  & 89.32           & $1\fs,H0$            & 91.97 & Lower                    & 91.79           & \highest{92.83} \\
All wrist         & SVC                  & \highest{88.77} & $2\fs$               & 87.04 & Lower                    & 88.55           & 87.12           \\
All               & SVC                  & 92.91           & $2\fs,H1$            & 92.95 & Both                    & \highest{94.46} & 92.28           \\ \bottomrule
\end{tabular}}
\caption{Binary classification problem accuracies for WESAD.}
\label{Wesad_LOSOCV_binary_table}
\end{table}

We obtained the highest accuracies when features from all persistence diagrams used together in both ternary and binary classification tasks. However, not every persistence diagram contributed equally to the accuracy. For instance, if we used only the upper level set persistence of all physiological signals, we would end up with $61$ features (a much smaller set of features than the one used in the original study) yet having reasonably high accuracies (79.61\% and 92.47\%) for the ternary and binary tasks.

\begin{figure}[H]
\centering
\adjincludegraphics[height=.26\textwidth,Clip={0} {0} {0.23\width} {0} ]{differentWindowSize_Wesad_ternary.png}
\includegraphics[height = .26\textwidth]{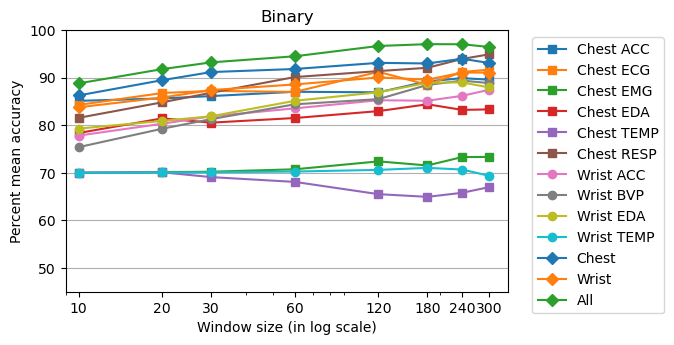}
\caption{WESAD accuracies for different window sizes.}
\label{differentWindowSize_Wesad}
\end{figure}

We expected higher accuracies when the window size gets larger. The subwindowing method allowed us to run the same learning algorithms for different sliding window sizes without extra computational cost. For this purpose, we rerun the learning algorithms using several window lengths ranging from 10 to 300 seconds. Our intuition turned out to be true: longer window sizes implied higher accuracy for most signals (Figure \ref{differentWindowSize_Wesad}). In particular, the accuracies for 300 second windows were as high as 89.86\% and 96.42\%, respectively for the ternary and binary tasks.

\begin{figure}
\centering
\includegraphics[width = .48\textwidth]{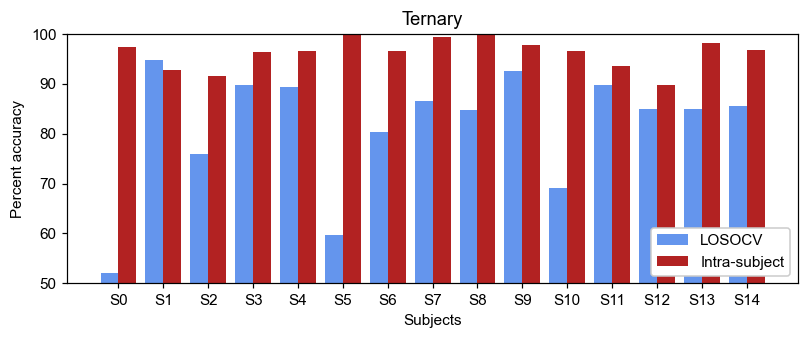}
\hspace{.2cm}
\includegraphics[width = .48\textwidth]{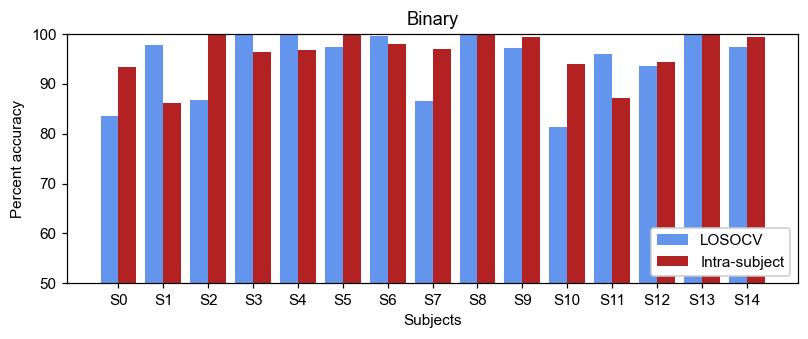}
\caption{LOSO and intra-subject cross-validation accuracies for WESAD.}
\label{Wesad_LOSOvsSameSubject}
\end{figure}

All of the analyses above were done using LOSOCV. A curious question at this point is to ask how much of these errors stem from individual differences. To answer this question, we used an intra-subject cross-validation.  We split the data for each subject in each condition into two subsets. Then we used one half to train and the other half to test the model, and \emph{vice versa}. The sliding window size was again chosen to be 60 seconds.  In Figure \ref{Wesad_LOSOvsSameSubject}, we compare these methods.

The difference was most pronounced in the ternary task. For instance, our intra-subject model performed 45\% better than our LOSO model. To test this effect, we ran two repeated measures \emph{t-}tests. For the ternary classification problem, our intra-subject model performed significantly better than our LOSO model. Our intra-subject model had an average accuracy of 96\%, while our LOSO model had 81\% with significance $p<.05$. Unfortunately, the accuracies for the binary classification task did not reach a statistical significance of $p < .05$ even though the mean accuracy was higher for the intra-subject rather than the LOSO model. This null finding is probably due to the ceiling effect. These findings indicate that when training and test sets contain data from the same subject, the learning algorithms perform significantly better.

\subsection{DriveDB dataset}

The cross-validation scheme of the original DriveDB study is different than ours. They used 300-second non-overlapping windows with a leave-one-out cross-validation scheme where the model is trained on all windows but one, and tested on the remaining. So, they mix windows from all subjects. Since we wanted to keep our method consistent across datasets, we again followed the same LOSO and intra-subject cross-validation methods we used for WESAD.

The accuracy for our ternary LOSO model was 85.81\% with an $F_1$-score of 79.68\% when we used 60s windows (Table \ref{Driver_LOSOCV_ternary_table}). Our binary LOSO model performed significantly better: 98.07\% accuracy with an $F_1$-score of 97.97\% (Table \ref{Driver_LOSOCV_binary_table}). A noteworthy observation is that the highest accuracies were obtained from the RESP signal. This indicates that the stress levels of drivers can be accurately measured using only one signal. This greatly reduces the number of features for the learning algorithms.

\begin{table}[H]
\centering
\resizebox{.7\textwidth}{!}{\begin{tabular}{@{}llclclc@{}}
\toprule
 &
  \multicolumn{1}{c}{} &
  \multicolumn{2}{c}{\begin{tabular}[c]{@{}c@{}}Delay\\ Embeddings\end{tabular}} &
  \multicolumn{2}{c}{\begin{tabular}[c]{@{}c@{}}Level\\ Sets\end{tabular}} &
   \\ \cmidrule(lr){3-4}
\cmidrule(lr){5-6} &
  \multicolumn{1}{c}{Clf} &
  Acc &
  \multicolumn{1}{c}{Dgm} &
  Acc &
  \multicolumn{1}{c}{Dgm} &
  \begin{tabular}[c]{@{}c@{}}All\\ Dgms\end{tabular} \\ \midrule
ECG      & RF  & 53.32           & $2\fs,H1$  & \highest{58.49} & Upper & 56.68           \\
EMG      & RF  & 66.87           & $.5\fs$    & \highest{69.14} & Upper & 66.49           \\
Foot GSR & RF  & 79.30           & $.5\fs$    & 79.60           & Both & \highest{80.08} \\
Hand GSR & RF  & \highest{67.26} & $2\fs,H0$  & 65.42           & Upper & 65.80           \\
HR       & SVC & 59.30           & $1\fs,H1$  & 59.80           & Both & \highest{60.61} \\
RESP     & SVC & 81.28           & $.5\fs$    & 85.71           & Both & \highest{85.81} \\ \midrule
All      & SVC & 82.50           & $.5\fs,H1$ & \highest{85.15} & Both & 80.59           \\ \bottomrule
\end{tabular}}
\caption{Ternary classification problem accuracies for DriveDB.}
\label{Driver_LOSOCV_ternary_table}
\end{table}

\begin{table}[H]
\centering
\resizebox{.7\textwidth}{!}{\begin{tabular}{@{}llclclc@{}}
\toprule
 &
  \multicolumn{1}{c}{} &
  \multicolumn{2}{c}{\begin{tabular}[c]{@{}c@{}}Delay\\ Embeddings\end{tabular}} &
  \multicolumn{2}{c}{\begin{tabular}[c]{@{}c@{}}Level\\ Sets\end{tabular}} &
   \\ \cmidrule(lr){3-4}
\cmidrule(lr){5-6} &
  \multicolumn{1}{c}{Clf} &
  Acc &
  \multicolumn{1}{c}{Dgm} &
  Acc &
  \multicolumn{1}{c}{Dgm} &
  \begin{tabular}[c]{@{}c@{}}All\\ Dgms\end{tabular} \\ \midrule
ECG      & RF  & 65.74           & $2\fs$     & \highest{71.11} & Upper & 67.61           \\
EMG      & RF  & 79.54           & $1\fs,H0$  & 82.04           & Both & 79.91           \\
Foot GSR & RF  & 90.83           & $.5\fs$    & \highest{92.26} & Upper & 91.88           \\
Hand GSR & RF  & \highest{82.47} & $1\fs,H0$  & 78.75           & Upper & 81.44           \\
HR       & SVC & 73.83           & $1\fs$     & 71.06           & Both & \highest{74.67} \\
RESP     & RF  & 93.27           & $1\fs,H0$  & \highest{98.07} & Both & 95.54           \\ \midrule
All      & RF  & 94.96           & $.5\fs,H1$ & \highest{96.24} & Upper & 95.39           \\ \bottomrule
\end{tabular}}
\caption{Binary classification problem accuracies for DriveDB.}
\label{Driver_LOSOCV_binary_table}
\end{table}

\begin{figure}[H]
\centering
\includegraphics[height = 4cm]{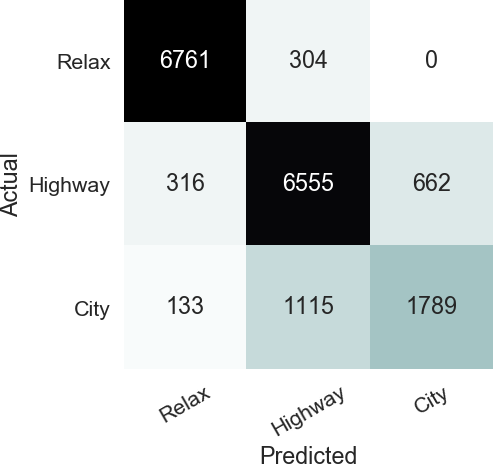}
\caption{Three-class problem confusion matrix for DriveDB.}
\label{Driver_confusion}
\end{figure}

Similar to the WESAD findings, we again found a separation between stress and non-stress conditions (Figure \ref{Driver_confusion}). The learning algorithms could easily distinguish \emph{relax} from \emph{city} and \emph{highway} conditions.

\begin{figure}[H]
\centering
\adjincludegraphics[height=.27\textwidth,Clip={0} {0} {0.23\width} {0} ]{differentWindowSize_Driver_ternary.png}
\includegraphics[height = .27\textwidth]{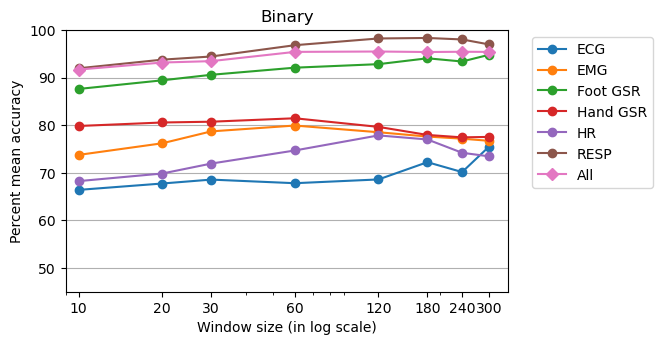}
\caption{DriveDB accuracies for different window sizes.}
\label{differentWindowSize_Driver}
\end{figure}

Also in this dataset, a greater window size had a positive effect on classification accuracies (Figure \ref{differentWindowSize_Driver}). When the window size is 300 seconds, our models had 91.19\% and 96.96\% accuracies for the 3- and 2-class tasks, respectively. 

\begin{figure}[H]
\centering
\includegraphics[width = .48\textwidth]{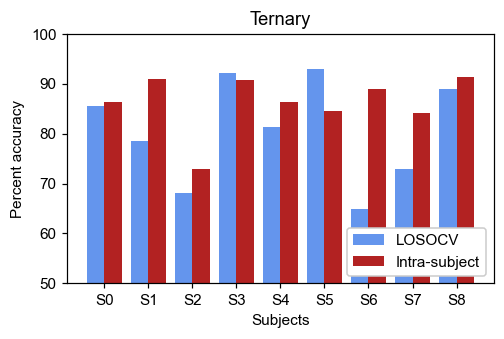}
\hspace{.2cm}
\includegraphics[width = .48\textwidth]{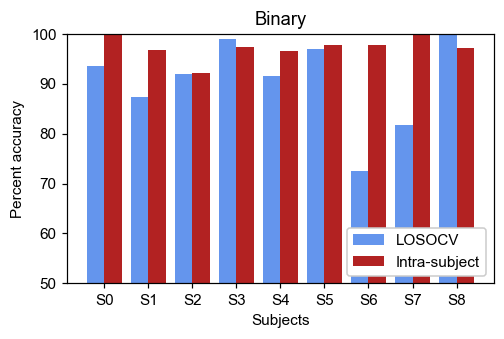}
\caption{LOSO and intra-subject cross-validation accuracies for DriveDB dataset.}
\label{Driver_LOSOvsSameSubject_ternary}
\end{figure}

Lastly, we tested if there are any differences in the accuracies of our intra-subject and LOSO models for 60s windows (Figure \ref{Driver_LOSOvsSameSubject_ternary}). To accomplish this task, we ran two repeated measures $t-$tests for the ternary and binary classification tasks separately. Although the mean accuracies were higher for intra-subject rather than LOSO cross-validation, the $t-$tests did not reach statistical significance of $p<.05$. We believe that the null findings are a consequence of small sample size and ceiling effect.

\section{Conclusion} \label{sect:conclusion}

The main aim of this study was to demonstrate the power of Topological Data Analysis (TDA) techniques in classification of time series. Specifically, we used persistence diagrams and their statistical properties to distinguish physiological signals collected under stress and non-stress conditions. As common in previous studies, this was done by creating sliding windows of a fixed duration, computing features, and training and testing machine learning models on these engineered features. The subwindowing approach we developed allowed us to inspect how the signal behaves locally, and how this local behavior varies over longer periods. Then, using TDA methods, we were able to create persistence diagrams from subwindows, create features on persistence diagrams and apply machine learning algorithms.

A proper feature engineering of a signal usually requires field knowledge. For instance, heart rate variability can be derived as a feature from an ECG signal. Our findings showed that most of the automatically generated topological features are at least as effective as signal-specific features in affect recognition. Under our models, the topological features we generated from the acceleration signals produced significantly better results than the baseline WESAD findings. Combined with the respiration signals, the accuracies of our models improved even further. Given that nearly all smartwatches and smartbands already have built-in accelerometer sensors, one can easily see that our method can readily, widely and cheaply can be used in other studies and applications.

{To test the effectiveness of our methods thoroughly, we also used a synthetic dataset in which elevated heart rate, elevated respiration rate, and variability in the heart rate were taken to be indicators of stress. On the synthetic dataset, our methods worked nearly perfectly in distinguishing stress from the baseline even in the presence of high noise. This tells us that the lower accuracies we obtained from the real datasets might be due to the limitations of our hypothesis on stress indicators we used for our synthetic dataset. It is possible that instead of a \emph{sustained} elevation, random sporadic changes in the heart and respiratory rate might be an accurate indicator of stress. Also, it is possible that the stress conditions in the experiments (e.g. driving in the city) may not cause observable stress in the subjects. Therefore, it is unrealistic to expect to see high accuracies from machine learning models on the real datasets we used in this study.}

The features in this study came from four different delay embeddings and level sets, which allowed us to make comparisons. The highest accuracies were obtained when we used features from all of the diagrams. Although this forces us to use a large number of features, it was possible to use features from only one delay embedding or level sets to get much fewer features, and obtain slightly less accuracy.

We have seen that longer windows usually implied higher recognition accuracies. However, this especially becomes computationally problematic when features are directly computed from the whole window. So, we implemented the subwindowing strategy, which was previously used \cite{honig2007fast} on a different dataset with different feature engineering methods. Using this method, one computes features on the subwindows, then finds the window features by taking averages and standard deviations. We observed that once the mean and standard deviation of a window is known, and when an observation is replaced with a new one within the window, there is a relatively cheaper computing strategy for the new mean and standard deviation regardless of the window size. Using this observation, we could use longer windows to compute new features from sliding windows and their subwindows to improve accuracy without incurring heavy computational costs.

Our findings also indicate that when the learning algorithm is trained and tested on disjoint data subsets coming from the same subject, the accuracies are higher than the LOSOCV approach. Hence, a device monitoring stress from a person may start with a relatively lower recognition accuracy, but it is going to improve its discriminative power when worn by the same person for long enough to update the parameters of the learning algorithm. In the original DriveDB study, the authors used 300 second non-overlapping windows with leave one out cross-validation (using both intra- and inter-subjects data) and reached approximately 97\% accuracy. When we trained our model on 300 second windows with LOSOCV, we reached higher than 91\% accuracy with fewer features. Given that we only included $9$ subjects (compared to $24$ in the original study) due to missing data and we used LOSOCV, our findings are promising.

\subsection{Future work}

This work used a single subwindow size for every stress condition in the signal. For future works, one can manipulate the subwindow size of a periodic signal and make sure that the delay embedding contains one and only one closed loop. Also, in addition to the $\varepsilon$ parameter in the Rips filtration, one can compute the multiparameter persistent homology where the dataset grows with time in the other dimension. The feature engineering methods can include other vector representations of persistence diagrams, and they can be combined with classical signal-specific features. Lastly, since no classification method appears to be a clear winner, one might consider using ensemble learners to increase the accuracy levels on these datasets.

\section{{Appendix}} \label{appendix_section}

This section contains the pseudocodes for the functions used in this study. Most functions can be parallelized. For instance, once the subwindows are created, one can compute the corresponding persistent homology in parallel for all subwindows.

\begin{algorithm}[H]
  \SetAlgoLined
  \DontPrintSemicolon
  \SetKwProg{Fn}{Function}{}{end}
  \KwIn{$x$: univariate time series. \linebreak $\fs$: sampling frequency. \linebreak $\ell$: subwindow length (in seconds). \linebreak $\kappa$: subwindow shift (in seconds).}
  \KwOut{$y$: the sequence of sliding subwindows}
  \Fn{getSubwindows}{
    Let $L$ be the length of $x$.\;
    $s\leftarrow \kappa \cdot \fs$\;
    $a\leftarrow 0$\;
    $b\leftarrow \ell \cdot \fs - 1$\;
    $i \leftarrow 0$\;
    \While{$b < L$}{
      $y_i \leftarrow (x_a,\ldots,x_b)$\;
      $a \leftarrow a + s$\;
      $b \leftarrow b + s$\;
      $i \leftarrow i+1$\;
    }
    \Return $y$}
\caption{The subwindowing algorithm.}
\label{subwindowing_algorithm}
\end{algorithm}

\hphantom{a}

\begin{algorithm}[H]
  \SetAlgoLined
  \DontPrintSemicolon
  \SetKwProg{Fn}{Function}{}{end}    
  \KwIn{$x$: subwindow. \linebreak $d$: embedding dimension.}
  \KwOut{The delay embedding of $x$ into $\mathbb R^d$.}
  \Fn{delayEmbedding}{
    Let $L$ be the length of $x$.\;
    \For{$i=0$ \KwTo $L-d$}{
      $y_i \leftarrow (x_i,\ldots,x_{i+d-1})$\;
    }
    \Return $y$}
\caption{Delay embedding of a subwindow.}
\label{delay_embedding_algorithm}
\end{algorithm}

\hphantom{a}


\begin{algorithm}[H]
\SetAlgoLined
\DontPrintSemicolon
\SetKwProg{Fn}{Function}{}{end}
\KwIn{A subwindow $sw$ with sampling frequency \fs.}
\KwOut{The sequence of persistence diagrams as in Figure \ref{subwindow_persistence_diagrams_pipeline}.}
\Fn{getDiagrams}{
  $D_1\leftarrow $ upper level set persistence of $sw$.\;
  $D_2\leftarrow $ lower level set persistence of $sw$.\;

  \For{$i$ = $1$ to $4$}{
    dataset $\leftarrow$ delayEmbedding($sw$, $i\cdot\fs/2$)\;
    $D_{2i+1}\leftarrow $ $H_0$-barcode of dataset.\;
    $D_{2i+2}\leftarrow $ $H_1$-barcode of dataset.\;
  }
\Return $(D_1,\ldots,D_{10})$}
\caption{Computation of persistent homology from subwindows.}
\label{persistent_homology_algorithm}
\end{algorithm}

\hphantom{a}

\begin{algorithm}[H]
  \SetAlgoLined
  \DontPrintSemicolon
  \SetKwProg{Fn}{Function}{}{end}
  \KwIn{A persistence diagram $D$}
  \KwOut{A 7-dimensional vector $f$}
  \Fn{getFeatures}{
    \For{$h\in D$}{
      $\ell_h\leftarrow death(h) - birth(h)$\;
    }
    $f_1 \leftarrow \frac{1}{\sqrt{2}}\|\ell\|_1$\;
    $f_2 \leftarrow \frac{1}{\sqrt{2}}\|\ell\|_\infty$\;
    $S \leftarrow \sum_{h\in D} \ell_h$\;
    $p \leftarrow \frac{1}{S} \ell$\;
    $f_3 \leftarrow - \sum_{h\in D} p_h\ln(p_h)$\;  
    
    Let $\beta$ be the Betti curve of $D$\;
    $f_4 \leftarrow \|\beta\|_1$\;
    $f_5 \leftarrow \|\beta\|_2$\;  
    
    Let $\lambda$ be the landscape vector of $D$\;
    $f_6 \leftarrow \|\lambda\|_1$\;
    $f_7 \leftarrow \|\lambda\|_2$\;
    \Return $(f_1,\ldots,f_7)$
  }
  \label{feature_engineering_algorithm}
  \caption{Feature engineering.}
\end{algorithm}

\hphantom{a}

\begin{algorithm}[H]
  \SetAlgoLined
  \DontPrintSemicolon
  \SetKwProg{Fn}{Function}{}{end}
  \KwIn{$x$: for a particular feature where $x_i$ is the feature value for $i$-th subwindow. \linebreak $M$: the number of subwindows in a window}
  \KwOut{Rolling mean and standard deviation of subwindow features.}
  \Fn{windowFeatures}{
    Let $N$ be the length of $x$\;
    $\mu_0\leftarrow \frac{1}{M}\sum_{i=1}^M x_i$\;
    $\sigma_0^2 \leftarrow  - \mu_0^2 + \frac{1}{M}\sum_{i=1}^M x_i^2$\;
    \For{i=1 \KwTo N-M}{
      $\mu_i \leftarrow \mu_{i-1} + \frac{1}{M}(x_{i+M-1} - x_{i-1})$\;
      $\sigma_{i}^2 \leftarrow \sigma_{i-1}^2  + \mu_{i-1}^2 - \mu_{i}^2 + \frac{1}{M}(x_{i+M-1}^2 - x_{i-1}^2)$\;
    }
    \Return $(\mu,\sigma)$\;
  }
  \caption{Rolling mean and standard deviation of features}
  \label{rolling_mean_algorithm}
\end{algorithm}

\hphantom{a}

\section*{Conflict of interest}
The authors declare no conflict of interest.

\section*{Funding}
This research did not receive any specific grant from funding agencies in the public, commercial, or not-for-profit sectors.

\bibliography{stressArticleReferences.bib}

\end{document}